\documentclass{article}
\usepackage{iclr2026_conference,times}
\iclrfinalcopy

\newif\ifanonymous \anonymousfalse
\newif\ifstatements \statementstrue
\newcommand{\releaseurl}{ (\url{https://github.com/digdoug/steering-dose-reparametrization})}
\newif\ifextended \extendedtrue
\newif\ifextapp \extapptrue

\usepackage{hyperref}
\usepackage{url}
\usepackage{booktabs}
\usepackage{longtable}
\usepackage{amsmath}
\usepackage{amssymb}
\usepackage{graphicx}
\usepackage{xcolor}

\graphicspath{{./}}

\newcommand{\dec}{\ensuremath{\mathrm{dEC}_{50}}}
\newcommand{\Lam}{\ensuremath{\Lambda}}

\ifanonymous
  \newcommand{\citeprior}{\citep{prior2026present}}
\else
  \newcommand{\citeprior}{\citep{prior2026presentnamed}}
\fi

\title{Activation Steering Transfer to Agents:\\ One Gain Ratio Does Not Identify Potency and Efficacy}

\author{Lucas Pinto\\Independent Researcher}

\begin{document}

\maketitle

\begin{abstract}
Additive activation steering is calibrated in single-turn chat and deployed inside agent
scaffolds. The quantity usually reported for that move is a \emph{gain}: a ratio of steered
effects, $T=\Delta_{\text{agent}}/\Delta_{\text{chat}}$. It is a summary of the same kind as a ratio of Braun's
steerability slopes \citep{braun2026steerability}, though not an identity with one: the two coincide
only under the proportional linear response this paper rejects. It is structurally a member of the
relative-steerability family of \citet{tan2024generalisation}. We show on eight family$\times$arm
dose--response cells over six models, swept in \emph{both} deployment contexts, that this ratio
\textbf{does not identify potency and efficacy}: reconstructing the published estimator in both of its forms on our own
grids, its realized range contains $1$ in five of five scorable cells, it moves with dose in four of
five, and two cells with opposite potency shifts, both CI-clean on the primary grid, return gain
intervals that overlap at a
width under $0.08$. Every scorable cell in the frozen five-cell gain audit is an amplifier at one dose
and an attenuator at another, so an amplify/attenuate taxonomy reports the dose it was read at.
We replace the gain with a \textbf{location}: $\dec = \mathrm{EC}_{50}^{\text{agent}} -
\mathrm{EC}_{50}^{\text{chat}}$, the cross-context difference in a curve location whose
context-dependent horizontal reparametrization \citet{taimeskhanov2026strength} derive analytically,
in a simplified model, for one context at a time. It is a curve location rather than a ratio of
effects, signed in both directions on this roster's primary grid across four distinct model families
($+1.013\,[+0.777,+1.273]$ against $-12.368$, $-10.855$, $-5.497$ and $-886.066$ elsewhere), and beats
a vertical rescaling at equal complexity in all five cells of the frozen horizontality audit
(four under the registered $\alpha^{*}$-trim).
Because separating a mechanism from an artifact is the question at issue, we report the
discipline at the same volume as the result: one cell is \textbf{quarantined loudly}, because its agent
response inverts past a coherence collapse and the fit's orientation flips, manufacturing the
roster's largest positive, which we delete. Three deflationary accounts (residual-norm rescaling,
baseline alignment, dose transduction) are measured and rejected on sign pattern and magnitude,
including \emph{wrong-sign} predictions on the roster's positive anchor. Our pre-registered forecaster
of $\dec$ was refuted out of sample and published as refuted, our registered salvage claim (a
location estimable where a gain ratio dies against a ceiling) produced no such cell and is reported
unanswered, and a census of our own pre-registration register reports the
registered branches no code in this repository could have emitted. The practical consequence is a measurement
instruction rather than a theorem: a transfer conclusion read at one strength does not identify what
changed, because a displacement and a gain are not distinguishable from a single operating point.

\end{abstract}

\section{Introduction}
\label{sec:intro}

Additive activation steering (extract a direction by contrasting trait-exhibiting and
trait-suppressing activations, then inject $\alpha \cdot u$ into the residual stream during
generation) is calibrated almost entirely on single-turn chat prompts. The models it is calibrated
on are deployed as tool-using agents: ReAct loops with format instructions, tool schemas,
observations and multi-turn transcripts. Whether a coefficient chosen in the first setting means
anything in the second is a deployment question with a safety edge.

It is also, and this is the paper's subject, an \textbf{identification} question that the field's
standard instrument does not answer. The quantities in wide use summarise a dose--response by a slope
over a fixed multiplier window \citep{tan2024generalisation,braun2026steerability} or by a win rate at
a chosen strength \citep{agarwal2026crossarch}. Neither holds position on the dose axis fixed. A recent
systematic measurement of chat-to-agent transfer \citeprior{} reports the resulting ratio
$T = \Delta_{\text{agent}}/\Delta_{\text{chat}}$ as a per-model constant (an amplifier at one end,
an attenuator at the other) and states in its own limitations that every $T$ was read at a single
dose, and that three of its families yielded no $T$ at all against a ceiling or a floor. A ratio of
two effects at one operating point is a property of the model only if the curves are proportional, and
nothing established that they were.

We sweep the whole curve in both contexts and ask what actually differs between them.

\paragraph{Contributions.}
\begin{enumerate}\itemsep2pt
\item \textbf{The gain ratio does not identify potency and efficacy} (\S\ref{sec:nonid}). Reconstructed in both of its published
forms and run on grids built to support a midpoint, its realized range contains $1$ in 5 of 5 cells, it
disagrees with itself across its own two forms in 5 of 5, and opposite-signed displacements return
overlapping gain intervals. Two of \emph{our} registered predictions were refuted in establishing
this, and both are reported.
\item \textbf{An identified estimand} (\S\ref{sec:estimand}), $\dec$ and its dose-unit-free companion
$\Lam$, read from curves swept until the fit's asymptote is
identified, which is not the same as every context reaching saturation and one does not, under a
registered grid rule
and a validity ceiling $\alpha^{*}$ that theory motivates and
we determine empirically.
\item \textbf{A location distribution, a roster and not a rate} (\S\ref{sec:law}):
8 cells, 6 models, 5 CI-clear sign-carrying cells across 4 distinct models, signed both ways on the primary grid; horizontal beats vertical by $\Delta$AIC in the
frozen five-cell audit;
fitted asymptotes agree to $0.012$ on the frozen five, and to $0.013$ once
Gemma is included, while midpoints move.
\item \textbf{Artifact discipline, at result volume} (\S\ref{sec:artifact}). One cell is quarantined,
at the cost of the roster's largest positive; two statistics of our own turn out to be artifacts of the
dose axis; and a census of our own register finds registered branches no code could emit.
\item \textbf{Three deflationary accounts, rejected on sign pattern and magnitude} (\S\ref{sec:defl}), plus
a fourth excluded by construction\ifextended{} and a chat-extraction \emph{transportability}
alternative probed by re-extracting the direction inside the agent's own frame, which bounds that
alternative on one model without establishing extraction invariance\fi.
\item \textbf{A registered predictor, refuted out of sample} (\S\ref{sec:predictor}), published as
refuted.
\end{enumerate}

\paragraph{What this paper does not claim.} Not that steering fails to transfer: the direction
installs, and the fitted asymptotes agree. We do not read that as equal achievable effect, for the
reason \S\ref{sec:estimand} gives. Not a validated forecaster of $\dec$. Not a causal
demonstration that editing the scaffold moves $\dec$: that experiment was designed, registered,
priced and \emph{not run} (\S\ref{sec:disc}). Not that any other paper measured the wrong thing. And
not that the two curves are translates of one another: our own test rejects that.

\section{Estimand, and what the theory motivates}
\label{sec:estimand}

\citet{taimeskhanov2026strength} prove three things about additive steering that \emph{motivate}
this design. We do not claim they force it: each is proved in a simplified model under its own
assumptions, and the steps from there to our estimand are ours. \textbf{Thm 3.6}: in a simplified model the concept-level curve is a $\tanh$ whose dose
axis carries a \emph{context-dependent horizontal reparametrization} $\nu_j$, derived in closed form
for \emph{one} context and never compared across contexts. Our estimand is the cross-context difference
in that curve's location, which on their form reads $\nu_j$ together with the baseline offset rather
than isolating it. \textbf{Thm 3.8}: for the training-set cross-entropy, and under
that theorem's assumptions, the first-order term of $\Delta\mathrm{CE}(\alpha)$ at $\alpha=0$ is
exactly zero and the leading behaviour is a variance-weighted quadratic. \emph{Our} inference from
it, not theirs, is that a susceptibility defined as a first derivative of any similarly sign-symmetric
aggregate will vanish to first order; our readouts are signed and directional for that reason. \textbf{Prop 4.1}: LayerNorm drives both contexts' \emph{raw} logits toward a shared
prompt-independent limit at large $|\alpha|$, collapsing the dynamic range any cross-context object
needs, though not, by itself, pinning a baseline-subtracted ratio to $1$. Either way it motivates a validity
ceiling $\alpha^{*}$ above which a reading is not trustworthy; we determine that ceiling empirically
rather than from the proposition, and \S\ref{sec:artifact} reports what it excludes.

The object to estimate is therefore a \emph{location on the dose axis}, with its fitted midpoints
inside $\alpha^{*}$, and with a signed readout:
\[
  \dec \;=\; m_{\text{agent}} - m_{\text{chat}},
  \qquad
  \Lam \;=\; \log m_{\text{agent}} - \log m_{\text{chat}},
\]
where $m$ is the fitted 4PL midpoint of the length-controlled combined binary response in each
context. We call this displacement a \emph{potency shift}, and reserve \emph{coherent efficacy} for
the second and separately measured dimension: whether output at that dose is still coherent.
$\Lam$ exists because a raw $\dec$ is in a \emph{model-specific} dose unit: our roster's
axes span $13$ to $4096$, so a raw cross-family magnitude comparison is a statement about axes rather
than about effects.

\paragraph{Why a location and not a gain.} A gain is a ratio of two response \emph{levels} at a
chosen dose and inherits that choice. A midpoint difference is a property of the whole curve, it is a
curve location of the kind Thm 3.6's reparametrization acts on, we read it signed for the reason
Thm 3.8's vanishing first-order term suggests, and (unlike a top-dose
ratio) it can be checked against the $\alpha^{*}$ that Prop 4.1 motivates.

\paragraph{Protocol and control spine.} Metric: a registered length-controlled \emph{combined}
binary, never judge-raw. $B=10\,000$ item-paired bootstrap, greedy decoding, $200$ item ids per
cell in both contexts and $N\ge 195$ in the resampled anchor. The \emph{fitted} set is smaller and
dose-dependent, since a rollout is dropped unless its metric is binary and its coherence flag is true;
the worst per-dose paired count among the law-eligible cells is $8$, on
\texttt{OLMo-2-7B:bypass}; the roster minimum is $0$, on the quarantined
\texttt{OLMo-2-7B:induce}. Controls: $\ge 5$ matched-norm random directions per
cell, entering as the per-dose $\Delta(\text{real}) - \Delta(\text{random})$ diagnostic and as the
admission gate rather than as a correction of the fitted response (App.~\ref{app:errata}); a
matched-information template ladder
from chat (C0) to a ReAct scaffold with deterministic dummy tools (C3), \emph{matched on the
instruction and asymmetric on the directive}: byte-identical instruction, but one span of the C0
system prompt is a compliance-valenced directive C3 does not
carry
\ifextapp{} (App.~\ref{app:m:spine})\fi\ifextended; and full KV recomputation at every turn,
verified in code rather than asserted, bounded by how often a rollout reaches the turns it covers
(\S\ref{app:m:spine})\fi. Four scope conditions are enforced in code and each can fail: \textbf{(i)} $\alpha^{*}$:
a cell is admitted only if both fitted midpoints sit
below the dose at which the real direction's cross-context separation collapses into a matched-norm
random band, contexts merged and read on the high side: the first dose above that separation's peak
at which the margin is non-positive; failing cells are marked \textbf{$\alpha^{*}$-RED}, printed, and excluded from
every law claim. \textbf{(ii)} the dose rule: $\ge 5$ doses spanning $\ge 4\times$ and bracketing
the midpoint, \emph{in both contexts}, or the cell is not fitted
\ifextended; of the $25$ two-context grids in the predecessor's whole raw corpus, exactly one
meets it, the cell it already published (App.~\ref{app:m:grids})\fi. \textbf{(iii)} a chat gate: a
$\ge 25$\,pt swing, or the family is dropped before it consumes a verdict. Its registered
\emph{``and beats matched random''} clause \textbf{never ran}; applying it now removes one cell,
which one is metric-dependent, and no sign-carrier.
\textbf{(iv)} the unit of analysis is the \emph{model}, never the cell.

\paragraph{Errata to the control record, carried rather than buried.} An audit of our own
control spine returns four record-level errata, most consequentially that the ``length-controlled''
metric is length-\emph{correlated}, holding on the field it names rather than on the property its
name asserts. All four are in Appendix~\ref{app:errata}; they correct the record; none of them changes the
estimand.

\section{The gain ratio does not identify potency and efficacy}
\label{sec:nonid}

\paragraph{First, what does survive.} The direction \emph{installs}: injected at the same layer and
coefficient, it arrives at the final layer in the agent context at install-site agent-over-chat ratios
of $0.83$--$1.16$ across three families \citeprior{}. And the two contexts' \emph{fitted upper
asymptotes} agree to within $0.012$ on the frozen five, $0.013$ once Gemma is included. We draw no efficacy conclusion from that: on a bypass cell the
asymptote is the floor refusal is driven to, so the contrast is blind to how much refusal there was to
remove, and \texttt{OLMo-2-7B:bypass} pairs a plateau contrast of $0.0000$ with a \emph{deliverable}
contrast of $-0.178$ whose CI excludes zero.

\paragraph{What the gain ratio does not determine.} We reconstruct the prior estimator in both of its
published forms (a single-dose ratio and an AUC-over-doses ratio, together with the per-dose
effect they are built from and the $\mathrm{SAT\_MARGIN}=0.05$ sub-saturation filter) copied
verbatim from its released analysis code,
and run it on our swept grids. Every per-dose effect recomputed reproduces the committed fit's own
record exactly ($\max|\Delta| = 0.0$, 7 of 7 cells), so what follows is a property of the estimator
and not of a re-implementation. This section, and the horizontal-vs-vertical and coherence tests of
\S\ref{sec:law}, were frozen on the seven-cell roster that predates the seventh family: their
denominators are the \textbf{five} $\alpha^{*}$-GREEN cells of that roster, against
Table~\ref{tab:core}'s six. Gemma-3-12B is law-eligible but post-dates that freeze; we name that fact
rather than re-open the audit.

\begin{table}[t]
\centering
\small
\caption{The scalar gain $T$ on grids built to support a midpoint. Every row is a property of the
\emph{estimator}, recomputed from rollouts that reproduce the committed fit exactly.}
\label{tab:nonid}
\begin{tabular}{@{}p{0.29\textwidth}p{0.65\textwidth}@{}}
\toprule
\textbf{finding} & \textbf{what it is made of} \\
\midrule
$T$ moves with dose & paired $\Delta T$ CI excludes $0$ in \textbf{4 of 5} $\alpha^{*}$-GREEN cells \\
$T$ spans $1$ everywhere & realized range contains $1$ in \textbf{5 of 5}: $0.067\!\to\!1.108$
(Llama-3.1-8B), $3.877\!\to\!0.590$ (OLMo-2-7B bypass), $2.011\!\to\!0.905$ (Qwen2.5-3B),
$2.607\!\to\!0.873$ (Qwen2.5-7B bypass), $10.050\!\to\!0.835$ (Qwen2.5-7B induce) \\
opposite displacements, overlapping gains & Llama at dose $9$ gives $T=1.0021\,[0.9908,1.0156]$;
Qwen2.5-7B induce at dose $53$ gives $T=1.0402\,[1.0047,1.0804]$: their potency shifts are
$+1.013$ and $-12.368$, \textbf{opposite signs, both CI-clean} on the primary grid (the $+1.013$
CI includes zero under the registered $\alpha^{*}$-trim, Appendix~\ref{app:astar-trim}), and the
gain intervals overlap at a
width under $\mathbf{0.08}$ \\
the estimator disagrees with itself & Llama is an \emph{attenuator} by the AUC form
($0.966\,[0.941,0.992]$, excluding $1$) and an \emph{amplifier} by the point form at its top dose
($1.108\,[1.060,1.166]$, excluding $1$): same rollouts, two estimators published in the same paper;
\textbf{5 of 5} cells fail both agreement clauses \\
\bottomrule
\end{tabular}
\end{table}

Table~\ref{tab:nonid} licenses one sentence, about the five cells it scores and not the RED cells or
Gemma-3-12B: \emph{every scorable cell in this frozen audit is an amplifier at one dose and an
attenuator at another}, so a per-model amplify/attenuate taxonomy reports the dose the ratio was read
at rather than a partition of models.

\paragraph{Two of our own registered predictions were refuted here, and both constrain what we may
say.} \textbf{(i)} On grids built to the dose rule, $T$ is computable in \emph{every} cell (34 of 48
cell$\times$dose pairs defined, no cell yielding none). The prior paper's three families that
``yield no coupling estimate'' are therefore a \textbf{grid} property, not a model property, and this
paper may not carry that sentence as a fact about those models. \textbf{(ii)} The \emph{identification level
set} (the (potency shift, efficacy ratio) pairs consistent with one measured gain at one dose, a
curve rather than a point) contains zero in only 2 of 3 scorable cells against a threshold of 3
registered in advance, so that half of our prediction fails.

\paragraph{What this section does not say.} $T$ is not \emph{wrong}: it is a well-defined functional
of two curves, the prior paper measured it at a stated operating point and said so, and the
relative-steerability family of \citet{tan2024generalisation} had established the form. Ours is the
narrower, identification claim: \textbf{a scalar read at one operating point does not determine the
pair (potency shift, coherent efficacy)}. A wide $T$ interval is not evidence \emph{for} $\dec$, which
enters here only as a label.

\section{The law}
\label{sec:law}

\begin{table}[t]
\centering
\small
\caption{Core result. $m$ is the fitted 4PL midpoint of the length-controlled combined binary in each
context; CIs are item-paired bootstrap, $B=10\,000$. $\alpha^{*}$-RED cells are reported in frame and
excluded from every law claim (\S\ref{sec:artifact}). Dose units are model-specific and are never
compared in magnitude across rows; see $\Lam$ in the text for the poolable axis.}
\label{tab:core}
\begin{tabular}{@{}llrrrlc@{}}
\toprule
cell & arm & $n$ & $m$ chat & $m$ agent & $\dec$ [95\,\% CI] & $\alpha^{*}$ verdict \\
\midrule
Llama-3.1-8B    & induce & 195 & 4.841    & 5.854    & $\mathbf{+1.013}$ $[+0.777,+1.273]$   & GREEN \\
OLMo-2-7B       & bypass & 200 & 17.875   & 12.378   & $\mathbf{-5.497}$ $[-7.851,-3.175]$   & GREEN \\
Qwen2.5-7B      & bypass & 200 & 17.049   & 6.194    & $\mathbf{-10.855}$ $[-15.814,-4.614]$ & GREEN \\
Qwen2.5-7B      & induce & 199 & 33.008   & 20.640   & $\mathbf{-12.368}$ $[-13.726,-10.876]$& GREEN \\
Gemma-3-12B     & induce & 199 & 2319.795 & 1433.729 & $\mathbf{-886.066}$ $[-945.445,-806.491]$ & GREEN \\
Qwen2.5-3B      & induce & 199 & 17.763   & 17.526   & $-0.237$ $[-1.303,+0.818]$            & GREEN \\
OLMo-2-7B       & induce & 199 & 16.695   & 37.114   & \textit{($+20.419$)} $[+18.863,+21.769]$ & \textbf{RED} \\
Starling-7B     & bypass & 200 & 7.307    & 6.942    & $-0.365$ $[-1.206,+0.488]$            & \textbf{RED} \\
\bottomrule
\end{tabular}

{\footnotesize\vspace{2pt}
\textbf{Committed $\to$ repaired 95\,\% CI} (the repair pre-registration requires both printed; the
committed interval is the one it dates, and no point estimate moves by more than $5\times10^{-4}$):
Llama-3.1-8B $[+0.777,+1.273]\to[+0.760,+1.295]$;
OLMo-2-7B:bypass $[-7.851,-3.175]\to[-7.687,-3.045]$;
Qwen2.5-7B:induce $[-13.726,-10.876]\to[-13.915,-11.017]$;
Qwen2.5-7B:bypass $[-15.814,-4.614]\to[-15.790,-4.253]$;
Qwen2.5-3B $[-1.303,+0.818]\to[-1.358,+0.818]$. CI-exclusion is unchanged on all five. Three rows
carry no repaired interval: two are $\alpha^{*}$-RED and outside the repair's frozen roster, and
Gemma-3-12B is law-eligible but post-dates it.\par}
\end{table}

\paragraph{Counts, at the registered unit.} Table~\ref{tab:core} holds \textbf{8 cells over 6 distinct
models}, \textbf{5 $\alpha^{*}$-GREEN models}, and \textbf{5 CI-clear sign-carrying cells across 4
distinct models} (Llama $+$; Qwen2.5-7B $-$, both arms; OLMo $-$; Gemma-3-12B $-$). We report the
model unit rather than the cell unit because Qwen2.5-7B enters as two arms of one model, and a
cell-unit count double-counts it.

\paragraph{Gemma-3-12B's $-886.066$ is never \emph{interpreted} without its scale-free
companion.} Its top dose is $4096$ against $13$--$77$
across the rest of the roster, a $315\times$ axis. On the log scale it reads $\Lam = -0.4812$,
within $0.012$ of Qwen2.5-7B induce's $-0.4695$: a $53\times$ larger top dose, a $71.6\times$ larger
raw $\dec$, one displacement. The sentence for that cell is \emph{``the agent context reaches
half-maximal effect at 62\,\% of the chat dose.''}

\paragraph{Heterogeneity, powered against per-sample variance.} \citet{gao2026cylindrical} hold that
steering sensitivity is controlled sample-specifically, so the aggregate has high per-sample variance
by construction; we power the family-level contrast against that variance rather than against cell
count. We report the specification our own pre-registration
requires, at the \textbf{model} unit over $\alpha^{*}$-GREEN cells only, since $\alpha^{*}$-RED cells
are declared not to contribute: $Q=712.51$ on $4$ df, $I^{2}=0.9944$, $\tau=9.432$ on the
$315\times$-span raw axis, against a design-minimum of $2.653$; joint sign power $0.9998$. $\dec$ is
family-dependent beyond per-sample noise, invariantly across all $14$ cells of the registered
specification lattice. Two facts bound it: Gemma alone supplies $93.0\,\%$ of $Q$, though deleting it
still leaves $Q=49.60$ on $3$ df and a powered verdict, and two of five leave-one-out refits return
\textsc{under\-powered}, so the verdict rests on the roster, not on any four of
its members. The axes being incommensurable, that pooling is a
heterogeneity statement and never an average effect; the same family of statistics computed on $\Lam$
($I^{2}=0.9886$ pooling cells, $0.9914$ pooling models) cross families; those are their own
pools and not the lattice cell above.

\paragraph{Horizontal, by test, and not further than the test supports.}
At equal model complexity a horizontal reparametrization beats a vertical rescaling in \textbf{5 of 5}
cells of that same frozen five-cell roster, every $\Delta$AIC interval excluding zero, and the
horizontality index $H$ (the
share of the context-induced gain in fit that a pure translation alone buys) has median $0.948$,
under a rule registered before the fit. \textbf{The registered $\alpha^{*}$-trimmed secondary is
weaker, and we report it as weaker}: $4$ of $5$, stratified by grid loss
(Appendix~\ref{app:astar-trim}). \textbf{But the curves are not translates}: writing $\mathrm{dEC}_q$ for the displacement read at
response level $q$, $\Delta_q = \mathrm{dEC}_{0.75} - \mathrm{dEC}_{0.25}$ is zero for a pure
translation and its CI \emph{excludes} zero in 3 of 5 cells, so
no sentence in this paper calls the two curves translates. The \emph{sign} is not an artifact of the
$0.5$ threshold: it survives five response levels (Llama is positive with a CI excluding zero at all
five, $\mathrm{dEC}_{0.10}=+1.247\,[0.94,1.82]$ through $\mathrm{dEC}_{0.90}=+0.779\,[0.33,1.20]$),
five link functions (25/25 cell$\times$link), and twelve fit windows (59/60); across an analytic
multiverse of 56 specifications the sign agrees in $\ge 0.9286$ of them in every GREEN cell. What does
\emph{not} survive is the functional form: $\mathrm{dEC}_{0.50}$ varies by up to $2.99\times$ across
links and the AIC-preferred link is our registered 4PL in only 1 of 5 cells, a prediction our own
pre-registration made \emph{against} our registered choice, and it held. The defensible claim is about
the sign.

\paragraph{Coherence is reported as an outcome and never used to exclude a cell.} Wherever the coherent margin can be decided it is
positive: 9 of 10 $\alpha^{*}$-GREEN cell$\times$context margins are established strictly positive, so
the midpoints we report sit \emph{below} the dose at which coherence half-collapses. That is what makes
them readable as behaviour rather than as decoding failure, and it is what the quarantined cell
violates.

\paragraph{The selection bound, signed and measured.}
\label{par:selection}
The positive side of ``two-sided'' rests on \textbf{one} model, and we quantify the exposure rather
than hedge it. All three agent-side measurability gates are monotone in absolute dose; the admission
window in 5 of 5 admitted cells reaches less far above zero than below it ($7.3\times$ harder upward), and the single admitted
positive sits at $88.5\,\%$ of its own upper edge. \textbf{The admission geometry therefore biases against observing
positive shifts}: the design under-observes exactly the sign that is scarce, so 1 in 5 is not
evidence that positives are rare, and five cells bound no population rate. The same cell carries a
rule qualifier: our estimator fits the full swept grid and reports $\alpha^{*}$ post hoc, not the
stricter dose-exclusion sentence two of our own pre-registrations carry. Re-run under that rule (same estimator, seed, $B$, items) the
positive cell reads $+1.352\,[-0.034,+1.587]$: \textbf{sign unchanged, CI-exclusion lost.}

\textbf{Deleting it would not cost the non-identification result}: $T$
still contains $1$ in $4$ of $4$ remaining cells and still moves with dose CI-cleanly in $3$, the
number we registered. It would cost the sharpest demonstration, since an opposite-signed pair needs
a positive shift to pair against.

\section{Artifact discipline}
\label{sec:artifact}

\begin{figure}[t]
\centering
\includegraphics[width=0.80\textwidth]{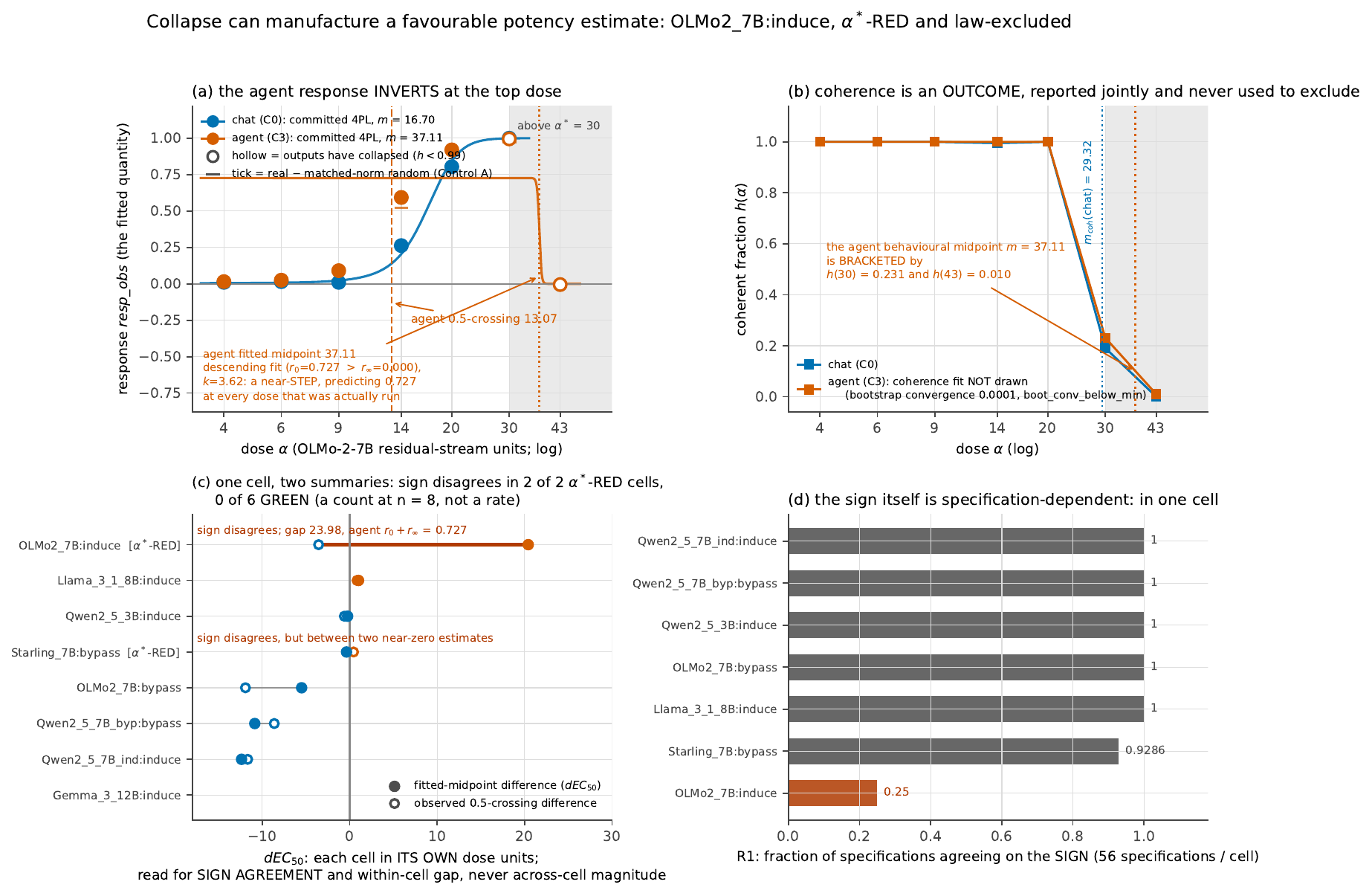}
\caption{\textbf{How a large positive ``shift'' is manufactured, and how the frozen screen caught it.}
One cell, \texttt{OLMo-2-7B:induce}; (c) and (d) add the roster's others: seven and six, the latter
roster frozen before Gemma-3-12B. \textbf{(a)} Behavioural dose--response in both contexts, on the
response the 4PL was fitted to, with the matched-norm random contrast overlaid as a tick rather than
subtracted (App.~\ref{app:errata}); marker fill
encodes the observed coherent fraction $h(\alpha)$, so a dose whose outputs have collapsed renders
hollow. The agent response rises to $0.995$ at dose $30$ and \emph{inverts} to $-0.005$ at dose $43$;
the fitted agent 4PL is consequently descending ($r_0=0.727 > r_\infty=0.000$) and places its midpoint
at $37.114$, past the collapse. \textbf{(b)} Coherence as an outcome: the agent coherence fit does not
converge (bootstrap convergence $0.0001$), so only its observed points are drawn. \textbf{(c)} Two
summaries of one \emph{cell}, not of one fit: the fitted midpoint $\dec$ against the model-free
$0.5$-crossing: $+20.419$ against $-3.556$, opposite in \emph{sign}, so this cell's shift is \textbf{not identified}. \textbf{(d)} Sign stability over 56 specifications. The cell is $\alpha^{*}$-\textbf{RED and law-excluded}, binding no fitted summary. Text at \S\ref{sec:artifact}; one cell, no prevalence claim.}
\label{fig:olmo}
\end{figure}

A venue asking what distinguishes a mechanism from an artifact is entitled to the cases where our own
numbers were the artifact. We report four, at result volume.

\paragraph{(1) The quarantine, and what it costs us.} \texttt{OLMo-2-7B:induce} is excluded from every
law claim. Figure~\ref{fig:olmo} shows why: past a coherence collapse the agent response inverts, the
fit's orientation flips, and its ``midpoint'' lands at no dose that was run, bracketed by
$h(30)=0.2312$ and $h(43)=0.0101$. The cell's fitted-midpoint summary ($+20.419$) and its
observed-crossing summary ($-3.556$) disagree in sign, and only $0.25$ of 56 specifications of this
cell agree on a sign against $\ge 0.9286$ everywhere else. \textbf{What the exclusion costs, said
plainly: the deleted cell is large and positive, so the quarantine removes the roster's biggest
positive and makes our two-sidedness claim harder to support.} Three readings are refused. The
crossing value is \emph{not} a corrected $\dec$: the cell is not identified. The cell is \emph{not}
a censored genuine positive; it cannot be both that and a collapse artifact. And this is \emph{not} a
critique of 4PL fitting: six cells in the same artifact are fitted by the same code at convergence
$1.0$, and the failure is specific to a grid carried past the collapse. The quarantine is at the
\textbf{arm}, not the family: \texttt{OLMo-2-7B:bypass} is $\alpha^{*}$-GREEN, CI-clear at $-5.497$,
and is one of the law's five sign-carriers.

\paragraph{A check on the screen that caught it.} Across the eight fitted cells the two summaries
disagree in sign in exactly the two $\alpha^{*}$-RED cells (2 of 2) and in none of the six GREEN
(0 of 6). $\alpha^{*}$ was frozen on an unrelated criterion long before anyone compared these
summaries, and the separation survived a test it could have failed when a seventh family added a GREEN
cell that does \emph{not} disagree. This describes a frozen screen post hoc rather than re-deriving
it, and $2/2$ and $0/6$ are counts at $n=8$, and we do not read them as rates.

\paragraph{(2) An explanation in a companion of ours that was a correlation with the dose axis.} Our
dose-resolution companion, released with the paper, explained two estimators' magnitude disagreements with a plateau-gap
correlation, $\rho = 0.893$ ($p=0.0068$) on seven cells. On a \emph{scale-free} axis (the same
disagreement as a fraction of the estimate rather than in raw dose units) those seven cells give
$\rho = 0.321$, $p = 0.48$. The correlation was with the dose axis, and one added model exposed it: on
eight cells the absolute-axis figure itself falls to $\rho = 0.571$, $p = 0.139$. The statistic was
non-binding (no verdict here rests on it) but it is a worked example of the failure this venue
names, found in our own companion, and the remedy is mechanical: \textbf{report every cross-family
diagnostic on a scale-free axis, or expect it to measure the axis.}

\paragraph{(3) A normalisation that succeeds and still does not deliver commensurability.} Because raw
$\dec$ lives in a model-specific unit, we asked whether dividing each cell's chat midpoint by that
model's chat residual norm makes the dose axis commensurable. It is \emph{highly effective and still
insufficient}, and both halves are measurements. Raw midpoints span $479\times$ across the six
$\alpha^{*}$-GREEN cells and normalised ones $21.7\times$, removing about 96\,\% of the cross-model
axis, while missing our registered commensurability threshold of $10$: on the five cells frozen
before the seventh family the normalised spread is $2.68$ and the claim holds; the sixth takes it to
$21.74$ and it fails. We report the reduction, which holds on all six; the commensurability claim holds on
five and we say so. The failure is not an instrument artifact: the sixth cell's norm leg is bound at the same layer
as its harvest, on the same 200 probes, at $5\,\%$ sd/mean, with a norm-ratio CI excluding $1$ like
every other cell's. The read is normal; the scale is not.

\paragraph{(4) A registered alternative that no code in this repository could produce.} A
pre-registration buys nothing for a clause whose losing branch cannot be emitted: that rule was
one-sided when it was scored, whatever the document says. We measured this rather than asserting it.
A census resolves all $299$ registered decision rules ($641$ branch mentions, $569$ distinct
labels) against every string the analysis code in this repository can produce, excluding by name the $41$
labels that are ordinary words, where a match is evidence of nothing either way. It found $4$
branches, both alternatives of two registered rules on a study that had \emph{landed}, with no emitter
at all, invisible because those rows sat behind a status the register validates by requiring the
artifact to be \emph{absent}. Both estimators are now written and both registered
predictions hit, though they were written after the analyst had seen the data and so are not blind
scoring. The $8$ landed branches that remain unemittable are each adjudicated by name in
Appendix~\ref{app:registers}, where the detector is also validated against $170$
demonstrably-emitted branches before any of its negatives is quoted. An earlier, more permissive
version turned all $40$ then-unemittable branches emittable and reported a clean tree.
\textbf{A detector that flags every branch cannot separate the emittable from the rest.} A second register
scored our frozen Methods draft against the pre-registration corpus and found it describing $10$ of $44$
registered designs (the corpus as it stood when the cut was made) and that gap is closed there, at
$45$ of $45$.

\section{Three deflationary accounts, measured and rejected}
\label{sec:defl}

Each of the three below is a mechanism a reviewer can name in one sentence, each is measurable
\emph{before generation}, and each is rejected on more than ``no effect'': on the sign pattern it
predicts, on the magnitude it can carry, and (on the roster's positive anchor) on \emph{pointing
the wrong way}. The rejecting evidence differs by account, and one of the three is mixed.

\paragraph{Residual-norm rescaling.} \emph{The objection:} the agent prompt has a different residual
norm, so a fixed $\alpha$ is a different relative perturbation, and the midpoint moves for a pure scale
reason. \emph{The measurement:} the norm ratio $\rho = \lVert r\rVert_{\text{agent}}/\lVert
r\rVert_{\text{chat}}$ is $0.86$--$0.91$ with a CI excluding $1$ in 5 of 5 cells: the effect is real,
and it is the other way round, the agent residual being \emph{smaller}. But $\log\rho$ is negative in
5 of 5, i.e.\ \textbf{sign-homogeneous}, while $\Lam$ is two-sided: $\Lam = +0.190$ on Llama against
$\log\rho = -0.118$ there, and negative-on-negative elsewhere. \textbf{A scale account predicts one
sign for every model; the data show two.}

\paragraph{Baseline alignment.} \emph{The objection:} the unsteered residual already has a component
along the steering direction, and that component differs by context, so what looks like a potency shift
is an offset. \emph{The measurement} (registered before the analysis code existed and before any such
cosine had been computed in this project): the cosine $c_0$ is nonzero in 5 of 5 cells (e.g.\ $-0.328$
chat vs.\ $-0.175$ agent on Llama) and context-dependent in 5 of 5 (\emph{the mechanism is real,
large, and present everywhere we looked}) and the composed scale-plus-translation law it generates
is nonetheless rejected in \textbf{3 of 5 scorable cells against a rejection threshold of 3 registered
in advance}. On the roster's only positive cell it predicts $\Lam_{\text{pred}} = -0.506$ against a
measured $+0.190$ (\textbf{the wrong sign}) and dividing the drive out makes the displacement
$3.7\times$ \emph{larger} ($+0.696\,[+0.628,+0.776]$). The affine scale factor of
\citet{cheng2026affine}, reported here because no one else has for this context pair, is inadequate at
median $R^{2}=0.189$. Two disclosures: our own registered prediction that the law would
\emph{undershoot} was refuted (the realized branch is mixed: 2 cells oppose in sign, 2 undershoot, 1
overshoots), and every $c_0$ here is read at the final prompt token while the deployment operator adds
$\alpha\cdot u$ at every position.

\paragraph{Dose transduction.} \emph{The objection:} $\dec$ is a difference of midpoints in
\emph{commanded-coefficient} units, and there is no reason a commanded unit buys the same residual
displacement in a 188-token ReAct prompt as in a 44-token chat prompt. \emph{The measurement:} the
objection is real. Delivered displacement per commanded unit differs by $+19.3\,\%$ (Llama) and
$-2.5\,\%$ (Qwen2.5-7B bypass), both intervals excluding $1$, \emph{and it runs the wrong way}.
Delivery predicts a $\dec$ of the \emph{opposite} sign on 2 of 3 cells with intervals excluding zero on
the wrong side, and explains $+3.0\,\%$ of the shift on the one cell where the sign agrees. Re-expressed
in chat-delivered units, all three cells keep sign and CI-exclusion and the only positive cell more than
doubles ($+1.013 \to +2.141\,[+1.883,+2.403]$). \textbf{Bound to its design}: three cells over two
models, one dose each, delivery and behaviour on \emph{independent} pools so it is not paired, and
normal-approximation midpoints.

\paragraph{A fourth account, excluded by construction rather than rejected by measurement.}
\citet{kang2026duality} attribute multi-turn coherence failure to KV-cache contamination. Every
agent-context rollout behind every $\dec$ here runs under mandatory full re-encode at every turn, so
cache reuse is absent \emph{by construction}, verified structurally in both agent code paths, a
check that can fail. Three scope limits bind. We have \emph{not} measured what cache reuse
contributes, only that our result does not depend on it; we do \emph{not} say
\citet{kang2026duality} are wrong, since the control they name is the one our protocol runs by
default; and recompute guarantees turn-uniform re-application, \emph{not} an exclusion of the duality
reading, our steering being turn-constant.

\ifextended
\paragraph{And the transportability alternative: that the direction is simply chat-native.}
\emph{The objection, and it is mechanical:} the direction was extracted in chat, so of course it
behaves differently under a ReAct scaffold. It is not a confounder --- the extraction frame was held
fixed within every paired deployment comparison we report, so it cannot be a common cause of context
and outcome. It is an \emph{effect-modification} alternative: it says our contrast would not
transport to a direction extracted elsewhere. \emph{The control:} a pre-registered
reciprocal-extraction matrix, in which the same instructions are also rendered inside the agent
frame. It is a horizontal sensitivity analysis at the fixed-effect level, \emph{not} a re-reading of
$\dec$: no 4PL, one or two real doses, the non-canonical parser metric, four agent turns. All 20
admissible cells, both frames and both arms, carry the canonical negative sign and every one excludes
$0$, and the maximum frame ratio is $1.494$ against a registered $2.0$.

\emph{On the induce arm the agent-native direction moves the displacement further from zero, and the
size of that move is a coordinate choice rather than a measurement.} We report it that way because
the arithmetic forces us to. All three admissible induce levels sit in one ladder segment, so the
paired frame contrast $G$ is \emph{exactly} affine in the ladder level --- $G = -0.226 - 32.42\,e$,
with a residual of $0$ --- and the ladder stops at $e = 0.15$ only because the chat-frame cell at
$e = 0.20$ returns a bootstrap-censoring fraction of $0.1195$ --- resampled draws with no bracketed
crossing, which is this instrument's own admissibility rule and not the coherence-collapse endpoint
of \S\ref{sec:artifact} --- against an inherited cut of $0.10$. Missing that cut by under two points
is what fixes the largest number we could have quoted: the same line reads
$-6.71$ at the excluded rung. So the finding here is the \emph{sign at every admissible level}
($3$ of $3$ exclude $0$) and the \emph{slope}, never a maximum. The dose-unit-free contrast behaves
as the deflationary reading of that would predict and the raw one does not: across the same ladder
$|G|$ ranges by $2.76\times$ while the log-potency contrast $G_{\Lambda}$ holds to $1.05\times$
($-0.827$ to $-0.871$, all three excluding $0$), and on the bypass arm $G_{\Lambda}$ is constant to
machine precision across a ladder over which $|G|$ ranges $6\times$. $G_{\Lambda}$ is also the only
one of the two that is \emph{not} an exact line in the level. Read on that scale, our chat-frame
direction understates the reparametrization on the induce arm by a factor of roughly $2.3$ in the
potency ratio itself, which a uniformly stronger direction would have left untouched.

That last reading is the narrowest in the paper: one model, one layer, one metric, and \textbf{one
independent block}, so the twenty cells are four blocks rather than twenty draws and we claim no
extraction-invariance from it. One arm is not favourable and is not written as one: on the bypass arm
the point estimates run the other way and \textbf{0 of 6} cells is CI-backed, so the frame question is
unresolved on that arm.

The two transport terms that could have made this a representation-geometry fact (an unsteered
offset and a transduction gain, both real, both measurable before generation) together account for
at most a quarter of $\dec$, with the wrong sign on the law's only positive cell.

\fi

\section{A registered predictor, refuted out of sample}
\label{sec:predictor}

\paragraph{The registration.} We frame this as what it is: \textbf{we learned the predictor is
wrong}. We do not present it as a weak positive. Before the seventh family's sweep existed, our pre-registration staked a
signed out-of-sample prediction on Gemma-3-12B: $\dec$ is \emph{positive}, and its normalised value
ranks 1st of 6, above Llama's $+0.0779$; refuted if the fitted $\dec$ is negative with a CI excluding
$0$, \emph{or} if its normalised value ranks below Llama's.

\paragraph{The outcome: both clauses fail.} Measured $\dec = -886.066\,[-945.445,-806.491]$: negative,
CI excludes $0$. Normalised, $-0.2163$ against Llama's $+0.0779$: not rank 1, not above Llama. The cell
is $\alpha^{*}$-GREEN, so no not-applicable escape was available and none was taken. The in-sample
Spearman correlations that motivated the feature were $+1.000$ at $N=4$ and $+1.000$ at $N=5$; they did
not survive one out-of-sample family. And \textbf{the map that fit best in-sample is the one that got the
sign wrong}: the frozen affine map forecast $+56\,245$ (a flagged extrapolation $6318\times$ outside
its training range) where a sign-only theory rule with no fitted parameter forecast negative. The
realized affine slope is $a = -0.0316\,[-0.0362,-0.0270]$ against the theory-implied $+1$: the
best-fitting map's own coefficients refute its own derivation.

\ifextended
\paragraph{The in-sample scorecard, for completeness and not as a rescue.} Leave-one-family-out,
no per-model fitted parameter, $N=4$ families on the common head-to-head set: an intercept-only
predictor gives LOFO MAE $9.39\,[8.06,10.84]$, $R^{2}_{\text{pred}}=0$; the training-free logit-lens
accuracy $A_{\mathrm{lin}}$ of \citet{billa2026where} gives $68.83\,[58.76,78.86]$ and
$R^{2}_{\text{pred}}=-312.9$; that paper's perturbation sensitivity $\lambda$ gives
$2.96\,[0.65,7.16]$ and $R^{2}_{\text{pred}}=0.908$; ours gives $0.282\,[0.230,4.864]$ and
$R^{2}_{\text{pred}}=0.9989$ at gauge \texttt{G0\_RAW} ($\kappa=0$). \textbf{Our registered verdict on that table was a stop, not a pass}: a
published within-context diagnostic \emph{also} recovers $\dec$ on this set, its MAE interval overlaps
ours, and at $N=4$ a leave-one-out affine map on four points is not identified against a collinear
competitor. We do not describe our predictor as validated, and the out-of-sample refutation settles it.
Sign accuracy is not a discriminator here: the head-to-head set is entirely negative-signed, so a
constant-sign rule scores $1.00$ by construction. \textbf{That last pair is
gauge-bound}: it holds at \texttt{G0\_RAW} ($\kappa=0$) and nowhere else, since every gauge the
registered audit calls defensible puts $R^{2}_{\text{pred}}$ below zero and the slope's sign moves
with the gauge.
\fi

\paragraph{One arm is a clean transfer failure.} At $R^{2}_{\text{pred}} = -312.9$, $A_{\mathrm{lin}}$
is $313\times$ worse than predicting the roster mean: a within-context steering-success diagnostic
predicts a cross-context shift \emph{worse in squared error than the intercept baseline}. The defensible verb is \emph{does not
transfer to} $\dec$, never \emph{refuted}: neither \citet{fan2026steerable} nor
\citet{billa2026where} claims a cross-context quantity, so this falsifies nothing they assert. The
per-example geometry arm of \citet{fan2026steerable} is \textbf{not scored} here: its classifier
needs per-example success labels this roster never produced.

\paragraph{Why $N=4$ and not 5, and why that is the same problem as \S\ref{sec:law}.} Llama-3.1-8B is
predictor-\emph{undefined}: its agent-side signed susceptibility sits at its own noise floor
($z=1.83$ against $z_{\text{crit}}=1.96$, versus $z=34.08$ in chat), and Llama is the roster's only
admitted positive. That set is therefore entirely negative-signed, the configuration in which a
sign-insensitive baseline is hardest to distinguish from a signed one: the selection geometry of
\S\ref{sec:law} and this negative are one problem surfacing twice.

\section{Discussion}
\label{sec:disc}

\paragraph{What it means for anyone steering an agent.} A coefficient calibrated in chat does not
carry a calibrated \emph{meaning} into an agent scaffold: the vector still works, but the same
behavioural fraction now sits at a different dose, in a direction not constant across models. Sweep
the curve in the deployment context as well as the calibration one.

\paragraph{An implication for how transfer is measured.} Neither
estimator family named in \S\ref{sec:intro} holds position on the dose axis fixed.
\citet{agarwal2026crossarch} reports a $71.0\,\%$ win rate at the best positive strength per pair
against a mean $\Delta = -0.004$ at a fixed strength for the same method, with 4 to 1 admissible
strengths per (model, method) by that paper's own usability threshold. \textbf{We do not decompose
that gap and claim no reading of it}: per-cell variances are unpublished, a maximum over $k$ noisy
estimates is upward-biased by selection alone, and the restriction is disclosed there.

\paragraph{The experiment we did not run.} The intervention that would establish $\dec$ as causally
\emph{movable} (editing the ReAct format scaffold with task information held fixed) was designed,
pre-registered, priced, and \textbf{not launched}. Its free forward-pass gate returned zero rankable
cells in 6 of 6 (family $\times$ readout-horizon) cells, so the planner refused to launch and no money
was spent. That refusal bought two reported things: a content-free placebo edit classifies the blocking
gate's two failure modes 6 of 6, separating ``reference set deficient'' from ``real information
lost''; and an attainability calculus priced the headline experiment out \emph{before} paying for it.
The readout invalidity is a property of this roster's 2024-era models, and we do not generalise it to ReAct scaffolds. On the pre-registered reading this is therefore a \emph{measurement} paper, and we say so rather
than reporting the shortfall as a result.

\paragraph{Limitations.} Beyond those stated where they arise: \textbf{(1)} the positive side rests on one model and a monotone
filter biases admission against positives, so the rate is not a population rate (\S\ref{sec:law});
\textbf{(2)} that cell's CI-exclusion is rule-dependent (\S\ref{sec:nonid}); \textbf{(3)} two nuisance parameters were chosen on one leg, only the
second swept\ifextended{} (\S\ref{sec:defl})\fi; \textbf{(4)} $c_0$ is read at the final prompt token while the operator acts everywhere, and
prompt-block variance understates generation-time variance; \textbf{(5)} a dose is not a shared unit, so every cross-family magnitude carries a
normalisation (\S\ref{sec:artifact});
\textbf{(6)} the curves are not translates, so $\dec$ summarises location at one response level
and the claim is about sign, not form; \textbf{(7)} four control-record errata are named
(\S\ref{sec:estimand}); \textbf{(8)} the causal half was registered, priced and not run; and \textbf{(9)} the register census bounds name-reachability, not
executability (Appendix~\ref{app:registers}). \ifextended{} The identification claim is least exposed to them: it is structural, not an
estimate (\S\ref{sec:nonid}).\fi
\label{body:end}

\ifstatements

\clearpage
\section*{Reproducibility statement}

Every number in this paper is transcribed from a committed machine-readable artifact and is
recomputed from those artifacts by code released with it. We separate that from two stronger claims we
do \emph{not} make. Regenerating the committed artifacts themselves from the released tree is only
partly possible, because the fits, control-spine audits and forecasts read rollout records and
activation bundles too large to ship; and regenerating the original model outputs is not possible from
the release at all, since it requires the inference runs themselves. Appendix~\ref{app:register} states
which of the three each artifact supports.
The release\releaseurl{} carries the fitting and scoring code, the frozen roster with pinned model revisions, the
pre-registration register of all 47 registered designs, the guard suite that checks the instruments,
and a self-test that runs inside the released tree. Appendix~\ref{app:apparatus} states which registered
branches were scored, which were not, and why.

Two limits are worth stating plainly. First, the scored rollouts behind the headline table are not
distributed in full: a majority of their bytes are model generations from a refusal-bypass arm, and we
withhold those for the reason given in the ethics statement below. The release instead carries the
same rollouts with the generated text removed, and the headline table refits identically from them,
including every midpoint, every bootstrap interval, and the heterogeneity and power blocks. Second,
the dose grids are model-specific and were frozen per family before any agent-side outcome was
observed, so a reader reproducing a cell must use the roster's grid for that cell rather than a
common one.

\section*{Ethics statement}

This work measures how much of an activation-steering intervention is required to change behaviour,
and one of the two arms it measures is refusal bypass. Studying that arm is what makes the paper's
safety claim possible: an intervention calibrated in single-turn chat does not carry its calibration
into an agent scaffold, so a control validated in one deployment context can be substantially
mis-dosed in another. Reporting that requires measuring the arm where the failure has consequences.

The corresponding risk is that the rollouts contain model outputs produced under a bypass
intervention. We therefore withhold those generations from the public release rather than publishing
them, and we do so without trading away reproducibility: the released rollouts retain every field the
estimators read and drop only the generated text, and the headline table refits identically from them.
No new jailbreak method is introduced here, and the steering technique itself is the standard
difference-of-means construction already in the literature we cite. All models used are publicly
released checkpoints at pinned revisions; no human subjects and no private or personal data are
involved.

\section*{AI use statement}

We used generative AI tools extensively in this work, and we state the extent precisely because the
disclosure is load-bearing for how the results should be read.

Of the tasks ICLR marks as requiring disclosure, we used AI tools to propose and refine hypotheses,
to design research methodology and experiments and to give feedback on them, to implement methods, to
clean and reformat data, and to interpret results against pre-registered criteria. We did not use them
for translation, and we did not use them to generate synthetic data sets: every number here comes from
rollouts of the pinned checkpoints. Support for qualitative and thematic data analysis does not arise;
all analysis reported here is quantitative. We did not develop theoretical models with them, did not use them to formulate mathematical
claims, did not use them to provide critical ingredients for proving mathematical claims, and did not
write proofs with them; the theoretical results we rely on are cited from prior work and are not our
own contributions. Of the tasks ICLR
marks as recommended for disclosure, we used AI tools for software implementation, literature search
and analysis, gap identification, figure generation, drafting and editing of this paper, reference
formatting, and title and keyword selection.

Concretely, the experimental programme was run by an autonomous research loop. That loop authored
pre-registrations, built the instruments and their tests, launched and monitored the compute, and
scored outcomes against registered branches. The human author set the research direction, chose which
questions to pursue and which to abandon, ruled on scope and cost, decided what the paper claims, and
reviewed the artifacts and the text.

We have reviewed all AI-assisted work, and the paper's verification apparatus exists because the
process was automated. Most studies were registered before their data existed, and the register is published with the
paper rather than summarised. It is not a universal claim and we do not make one: the control-spine
conformance audit is marked \textsc{post-data}, the causal-attainability derivation is marked
post-hoc on frozen artifacts, and the extraction-frame study is a pre-specified analysis of
already-public data rather than a blind pre-data registration. Each says so on its own face, and the
register records which kind each study is. The guard suite tests the instruments against
deliberately planted defects, so a check that cannot fail is ordinarily visible as such. That guarantee is not complete, and we state where it failed. The suite plants each defect against a check's \emph{declared} subject, so where a check's declared subject and its actual one diverge, the planted defect diverges with it. One rail here read a figure caption from that figure's provenance sidecar while its description, its inline comment and its mutant all said it read the caption the paper ships; rail and falsification test were aimed at the same wrong file, and the shipped caption had meanwhile lost the law-exclusion the figure exists to carry. Adversarial review found it and the suite did not. We report it for the same reason we report the refuted forecaster: an apparatus that records its own miss is worth more than one that claims completeness. A census of our own
register found registered branches that no code in the tree could emit, and we report the count and
each disposition in Appendix~\ref{app:apparatus} rather than repairing them silently. Our
pre-registered forecaster of the estimand was refuted on the first held-out family and is published as
refuted. These are not incidental: they are the reason we think results produced this way can be
checked by a reader who did not run them.

We take responsibility for the final content of this work, including all text, claims, and artifacts
produced with the aid of generative AI tools.

\fi

\appendix

\section{Errata to the control record}
\label{app:errata}

An audit of our own control spine returns four record-level errata. We state them rather than
silently repairing them, because each is a case where a control's \emph{name} asserted more than the
field it was computed on.

\begin{enumerate}\itemsep2pt
\item \textbf{Matched-norm random directions} enter as a per-dose diagnostic and as the $\alpha^{*}$
separability gate, rather than as a correction of the fitted response. Every midpoint in
Table~\ref{tab:core} is a baseline-corrected (induce) or absolute (bypass) fit; \textbf{none is a
random-corrected fit}, and no line of the estimator subtracts the random arm from the response the
curve is fitted to. Applied to the estimand as a reported sensitivity (same estimator, seed, $B$
and items) the correction leaves sign, CI-exclusion and status unchanged in 5 of 8 cells, and we
name the three it moves. Under the plain-reading \texttt{mean} aggregation \texttt{OLMo-2-7B:bypass}
strengthens from $-5.497$ to $-8.274$, \emph{away} from zero; under the conservative \texttt{maxabs}
band the same cell is censored, which would cost the law its fifth family. \texttt{Qwen2.5-3B:induce}
and the $\alpha^{*}$-RED \texttt{Starling-7B:bypass} change sign inside intervals that span zero in
the primary, and under \texttt{maxabs} alone \texttt{Qwen2.5-3B} reaches $+1.730\,[+0.328,+2.985]$, a
CI excluding zero. We read \texttt{maxabs} as a gate and never as an estimator (a maximum over five
draws is a positively biased estimate of a random direction's effect) but the registered primary is
the uncorrected fit either way, and that is the erratum.
\item \textbf{The ``length-controlled'' combined metric is length-\emph{correlated}}, not
length-controlling. It holds on the field it names (the combined binary is not the judge-raw score
that carried the length artifact) but the name asserts a property the construction does not
deliver, and a residual length--response correlation remains.
\item \textbf{One pairing control holds on the raw record and not on the set the estimator averages.}
Item pairing is exact in the logged rollouts; the coherence-gated subset the estimator averages over
can drop an item in one context and not the other at a single dose. The same gate also acts on a whole
\emph{dose}: each context's grid is derived downstream from the rollouts that survived scoring, so a dose
at which every rollout in one context has collapsed leaves that context's grid with no field recording
that it was ever swept, while the other context retains it and the two legs of a paired midpoint
difference are fitted over different ranges. We re-score the realised support of every cell in the
canonical fit on four rails: both contexts fitted on the same dose set; a finite bootstrap sd at every
fitted dose: the weighted fit floors a \emph{non-finite} sd to the same minimum it uses for a
saturated one, which hands the least-determined dose the greatest weight; no swept dose removed from a
fitted grid by the scoring filter; and the $\ge 5$ doses spanning $\ge 4\times$ rule re-checked on the
grid that was fitted rather than the one that was planned. All four rails pass on all 6 law-included cells
and all 12 legs, at a minimum scored $n$ of 26 per leg. Both conditions fire in exactly 1 cell, the
already-quarantined $\alpha^{*}$-RED \texttt{OLMo-2-7B:induce}: at dose 43 the chat context has 0 of
200 rollouts coherent and loses the dose, the agent context keeps 2 and retains it, and that retained
point's undefined bootstrap sd is floored to the minimum.
\item \textbf{The chat gate ran on one clause of two.} Its registered form is a conjunction (a
$\ge 25$\,pt swing \emph{and} beating matched-norm random); the executed check evaluated the swing
clause alone. The random comparison is made at the $\alpha^{*}$ stage, but as a \emph{cross-context}
separation against the random band rather than as the gate's per-family clause, so the registered
conjunction was never enforced in either place.
\end{enumerate}

None of the four moves a fitted number, a CI, or a verdict; they are errata to the record of how a
control was realised. Each is stated with what would falsify it in the full limitations
register\ifextapp~(\S\ref{app:limits})\fi.

\section{Pre-registration and audit apparatus}
\label{app:apparatus}

Every paid experiment in this project carries a dated pre-registration committed before the run it
governs, with its live/die branches (including its null and its adverse branches) written before
any number exists. Three machine-scored registers keep that discipline honest after the fact rather
than by vigilance: a \emph{pre-registration ledger} linking each registered rule to the verdict token
some artifact actually emitted; a \emph{methods register} that fails closed if a registered design's
procedure goes unstated in the full methods appendix or in the companion document its row names; and
a \emph{limitations register} that fails closed if a threat frozen before the data is absent from the
full limitations register.
\ifextapp{} Both scored texts are carried here
(\S\ref{app:methods}, \S\ref{app:register}, \S\ref{app:limits}).\fi{} A whole-manuscript consistency audit mechanically re-checks that every $\dec$ asserted in prose is a value the canonical fit contains and
that every ``CI excludes zero'' claim is true of the interval printed beside it. Two registered
verdicts reported in this paper (the predictor stop of \S\ref{sec:predictor} and the launch refusal
of \S\ref{sec:disc}) are branches those registrations wrote in advance and that we would have
preferred not to fire.

\subsection{Two audits of the apparatus, and what they returned}
\label{app:registers}

Both registers below were run on this repository and both returned defects. We report them at the
size the finding had, because a falsifiability claim that is never itself measured is exactly the
failure this paper is otherwise about.

\paragraph{Branch-emitter census.} For each registered branch label, is there a string in non-guard,
non-registry code that the tree can produce and that equals it: as a literal, an f-string with
substantial literal evidence, a run-time prefix rewrite, or a marker written by a shell launcher?
Scale: all $299$ registered decision rules, $641$ branch mentions, $569$ distinct labels, against
every string constant the analysis code in this repository can produce. Mentions resolve as \texttt{EMITTED\_IN\_CODE} $521$,
\texttt{NAME\_UNRESOLVABLE} $79$, \texttt{ONLY\_IN\_REGISTRY} $29$, \texttt{DERIVED\_IN\_CODE} $7$,
\texttt{COMPOSED\_IN\_CODE} $3$, \texttt{ONLY\_IN\_GUARD} $1$, \texttt{ONLY\_IN\_SHELL} $1$. The $79$
unresolvable mentions are the $41$ labels that are ordinary words (\texttt{PASS}, \texttt{NULL},
\texttt{GREEN}, \texttt{MIXED}, \dots); they are counted, listed and excluded from every verdict
rather than resolved in the direction that flatters the tree.

Three limits are declared. The census measures static \emph{name-reachability}, not executability: a
label found here is emittable in principle, and it does not prove the emitting line is reachable from
an entry point on the realized inputs. The error therefore runs in the safe direction: a label
\emph{not} found cannot have been emitted by name at all. And nothing here says whether a registered
prediction was any good; only whether the register could have recorded its failure.

\paragraph{What it found, and the disposition of what remains.} Twelve branches belonged to a rule
whose study had landed and had no emitter. Four were the real defect: both branches of two
registered rules on the cache-reuse study, an $\alpha^{*}$ recomputation and a grid diagnostic. Both
estimators were then written, both are refits of records that already exist, and both shipped the
branch their pre-registration had named in advance. The remaining eight each carry a disposition
naming the class it expects, so writing a missing estimator turns the entry stale and forces its
deletion: an explanation may not outlive the thing it explains.

\begin{center}\small
\begin{tabular}{lrl}
\toprule
disposition & $n$ & why it is not an owed estimator \\
\midrule
\texttt{STAGE\_NOT\_REACHED} & $3$ & Stage 1 was gated on a clean Stage 0; Stage 0 failed two of \\
 & & three bounds and Stage 1 was never launched, so the labels \\
 & & describe an experiment that does not exist in this tree \\
\texttt{RULE\_NOT\_EVALUABLE} & $3$ & the rule shipped its not-evaluable or withdrawn branch, so \\
 & & the measurement that would select between the substantive \\
 & & branches was never built \\
\texttt{GATE\_LABEL\_NOT\_A\_VERDICT} & $2$ & a named precondition, not a branch the scorer selects \\
 & & between \\
\bottomrule
\end{tabular}
\end{center}

A further $20$ branches belong to studies registered and not yet run, where no estimator is owed
yet. We state the number because it is the difference between \emph{we will run it} and \emph{we have
not written it either}, and it is part of what running those studies would cost.

\paragraph{Validating the detector before quoting its negatives.} A branch that is the realized
verdict of a rule scored in a committed artifact was demonstrably emitted, and the ground truth is
re-read \emph{from that artifact} rather than from the registry field being audited. There are $170$
demonstrably-emitted branches; the scanner must find an emitter for every one. Unwaived misses:
$0$. Waivers: $0$. Stale waivers: $0$. This is not ceremony. The first f-string rule required only one literal part, so
a two-variable template matched nearly every namespaced label and, in one run, all $40$
then-unemittable branches turned emittable and the census reported a clean tree. The repair was a
stricter detector, not a waiver; the waiver table exists, is subject to a stale-waiver rule, and is
empty.

\paragraph{Methods register.} A second register takes the mechanical set of pre-registration
documents as its denominator and scores the merged Methods document against the tree: a
\texttt{MAIN} row must name a section of that document that exists \emph{and whose own body cites the pre-registration}, so a row cannot
pass by being written down. Measured when the re-cut was made, against a corpus of $44$ pre-registrations,
the frozen Methods draft cited $10$ of $44$ registered designs by filename ($13$ by a looser stem
reading) and the
merged draft cited $5$ ($14$ by stem): $31$ registered designs (including every deflationary
measurement this paper calls its hardest-to-dismiss result) had no procedural description anywhere
a reviewer reads. The pre-registration ledger could not see it, because it asks whether every
registered \emph{rule} has a recorded \emph{verdict}, one step later in the reading order than
\emph{where is the procedure that produced this number}. The register now reports $47$ of $47$, $0$ defects,
disposed as $25$ stated in the full methods appendix, $18$ in the named companion documents,
$2$ infrastructure and $2$ registered-but-not-run.

\paragraph{The $\alpha^{*}$-trimmed horizontality secondary, per cell.}
\label{app:astar-trim}
Re-running the horizontality test on grids cut at the cell's single paired $\alpha^{*}$, applied to
both contexts, gives
\texttt{HORIZONTAL\_VALIDATED} again at median $H=0.964$, but $4$ of $5$ rather than $5$ of $5$.
Writing $\Delta\mathrm{AIC}_{R2}$ for the model-comparison margin (not a coefficient of
determination), full grid $\to$ $\alpha^{*}$-trimmed:
\texttt{Qwen2.5-7B:induce} $+347.87 \to +347.87\,[264.5,431.2]$ and \texttt{Qwen2.5-3B}
$+9.80 \to +9.80\,[1.4,22.2]$, both unchanged because neither carries an $\alpha^{*}$-excluded dose;
\texttt{Llama-3.1-8B} $+87.54 \to +38.45\,[16.6,70.7]$ on five of seven doses;
\texttt{Qwen2.5-7B:bypass} $+24.81 \to +0.17\,[-4.9,+4.9]$ on six of seven; and
\texttt{OLMo-2-7B:bypass} $+46.37 \to -11.87\,[-33.4,+6.6]$ on five of seven. The two zero-loss cells act as a control:
neither moves at all, which is what attributing the change to lost grid support requires. Among the
cells that do lose doses the ordering is \emph{not} simply how many go, since
\texttt{Qwen2.5-7B:bypass} loses one dose and collapses further than \texttt{Llama-3.1-8B} losing two.
\texttt{OLMo-2-7B:bypass}'s point estimate does reverse the preference; what we rely on is that its
interval spans zero, so the trimmed grid does not \emph{establish} a reversal, rather than that it
shows none. We report it because the primary fit
deliberately retains doses at and above $\alpha^{*}$, so a reader is owed the version that does not.



\section{Related work, in full}
\label{app:related}

The body concedes only what its guardrails require, in the space a 9-page limit leaves. This section
is the full positioning. Its organising claim is that \textbf{nothing here measures a ratio across two
deployment contexts for a fixed model and a fixed direction, and yet every measurement primitive this
paper uses is owned by someone else.} The job below is therefore to concede every primitive to its
owner at exactly the right altitude; under-conceding is the failure mode a reviewer punishes.

\paragraph{Why this section is in the appendix rather than the main text.} The 9-page body ends flush
with the bottom of its last page: it has no free line. Restoring this section to the main text would
cost a section of the identification argument, which is the contribution the body exists to carry. The
concessions the guardrails require (the slope-estimator credit, the kinship between $T$ and a ratio of
Braun's slopes $S$, the Billa altitude, the SVF phrase) are each discharged \emph{in the body} at their
first mention; what is here is the complete version, not the load-bearing part.

\subsection{The steerability slope and relative steerability}
\label{app:rel:tan}

\emph{Owned: the propensity-curve-and-slope machinery; steerability $s$; relative steerability
$s_{\text{rel}}$; and deployment-context manipulation at readout. We concede all four.}

\citet{tan2024generalisation} add $\lambda\cdot v_L$ at the last token position of layer $L$, read
propensity as the logit difference between correct and incorrect multiple-choice answer tokens, and
define \textbf{steerability} $s$ as the slope of a least-squares line fit to the mean scores at each
multiplier $\lambda_i$ over the grid $\Lambda=\{-1.5,-1.0,-0.5,0,0.5,1.0,1.5\}$. \textbf{Relative
steerability} $s_{\text{rel}}(v_A,D_B,\Lambda) = s(v_A,D_B,\Lambda)/s(v_B,D_B,\Lambda)$ is the ratio of
two such slopes measured in the \emph{same} application context. They also vary the deployment context
at readout, extracting under one prompt setting and applying under another.

Three concessions follow and we make all three explicitly. \textbf{(i)} The slope estimator is theirs;
where \S\ref{app:rel:braun} places $T$ alongside a ratio of Braun's slopes, Braun
borrows $S$ from here, and the credit for the primitive belongs at this citation. \textbf{(ii)} $T$ is
a member of the relative-steerability family (a ratio of two effect summaries formed to describe
transfer) so it was not a new object, and this paper says so at the first mention of $T$.
\textbf{(iii)} ``Steering vectors change effectiveness when the deployment context changes'' is not
ours to claim as new.

What is ours is the \emph{axis}, and it is a design difference rather than a correction. A slope taken
over $|\lambda|\le 1.5$ is not a location parameter: it does not separate a change in attainable effect
from a change in the dose that attains it. Two further separations are worth stating. Their context
shift injects instructions that name the steered trait, so the steerability change is entangled with a
shifted baseline propensity on the same axis, whereas a ReAct scaffold is behaviour-orthogonal to
refusal. And their grid carries three positive multipliers gated at the top, which fails this paper's
registered dose rule by a wide margin: even with their raw per-$\lambda$ data a midpoint could not be
fitted to it.

\paragraph{The obvious reinterpretation is dead, and we do not make it.} We do \textbf{not} write that
their conclusions are a multiplier-window artifact. They ran that ablation themselves and report that
the overall trends in steerability remain highly consistent across multiplier ranges. Making the claim
anyway would require arguing that their own registered robustness check is inadequate, using data we do
not have. Their readout compounds it: a logit difference selected partly for linearity, and a
horizontal shift of a locally linear curve does not change its slope.

\paragraph{One inherited overstatement we retire rather than repeat.} A secondary source summarises
this work as consistently finding lower effect sizes out of distribution. That is not what they report:
their headline is that steering vectors generalise reasonably well, and their principal scatter falls on
both sides of $x=y$ ($\rho = 0.891$ on Llama-2-7b, $0.694$ on Qwen-1.5-14b). Those are not to be
confused with their weaker panels ($0.535$/$0.341$; $0.769$ in-distribution and $0.586$ out;
$-0.26$/$-0.46$). We cite the weaker, correct version.

\subsection{The shape of the dose--response curve}
\label{app:rel:taim}

\emph{Owned: the shape of a single steering dose--response curve, and the three results that constrain
any susceptibility readout. Our estimand is the cross-context difference in the one quantity they
derive for a single context and never compare across contexts.}

\citet{taimeskhanov2026strength} give the closest theoretical account of what a steering dose--response
curve is, and three of their results are load-bearing here. Their \textbf{Thm 3.6} makes the
concept-level response a $\tanh$ up to a context-dependent horizontal reparametrization $\nu_j$, in a
simplified model: they derive $\nu_j$ in closed form (their App.~B.4) for a single context, and never
form a cross-context difference, ratio or comparison of it. Their \textbf{Thm 3.8} makes the first-order term of
$\Delta\mathrm{CE}(\alpha)$ exactly zero, for the training-set cross-entropy and under that
theorem's assumptions. \emph{We} read it as suggesting that a susceptibility defined as a first
derivative of any similarly sign-symmetric aggregate will vanish likewise; that step is ours, not
theirs, and our readouts are signed and directional because of it. Their \textbf{Prop 4.1} makes
LayerNorm force prompt-independence at large $|\alpha|$, so the contexts converge toward a shared limit
and the dynamic range that identifies any cross-context object collapses: a normalisation artifact.
It drives the \emph{raw} logits together; it does not by itself pin a baseline-subtracted ratio to
exactly $1$, since the contexts differ at $\alpha=0$. We therefore determine a single paired-context
$\alpha^{*}$ and report where the comparison sits relative to it. The fitted \emph{midpoints} lie
strictly inside $\alpha^{*}$ in both contexts; the swept grids do not, since the primary fit
deliberately retains doses at and above the ceiling and reports $\alpha^{*}$ post hoc. The registered
$\alpha^{*}$-trimmed secondary is reported in the body, and it is weaker than the primary.

The delta is one word: they derive \emph{one} curve's shape and its reparametrization analytically in a
simplified model; we estimate the difference in that curve's location \textbf{between two deployment
contexts}, empirically, with a CI. Their $\nu_j$ is a term inside our dependent variable, not identical
to it (\S\ref{app:lim:L2}).

\subsection{Per-example prediction of steering success within one context}
\label{app:rel:fanbilla}

\emph{Owned: early-decoding residual geometry, and cheap forward-pass diagnostics, as predictors of
steering outcome within a single context. Both are cited in the introduction. One is run as a
baseline; the per-example geometry arm is registered and \textbf{not scored}, for the reason the
Results give. This is where altitude discipline matters most.}

\citet{fan2026steerable} predict, within one fixed chat context, which (prompt, concept, strength)
triples steer successfully, from early-decoding residual geometry; their Steering Geometry feature group
is fully specified and is reused here as baseline P1. \citet{billa2026where} predicts \emph{where} (at
which layer, for which family) a steering vector will succeed, from training-free forward-pass
diagnostics: a logit-lens argmax accuracy $A_{\mathrm{lin}}$ and a perturbation sensitivity $\lambda$.

We take both as genuine prior art on their own question. Both of
\citeauthor{billa2026where}'s diagnostics are run as registered refutation arms. \citeauthor{fan2026steerable}'s
per-example Steering Geometry group, baseline P1, is registered and \textbf{not scored}: its classifier
needs per-example success labels this roster never produced.
The anchors we score them against are read from the published papers: $A_{\mathrm{lin}}$ predicts
steering effectiveness at $\rho = +0.86$ to $+0.91$ \emph{cross-model} across five models and layer
selection at $\rho = +0.63$ to $+0.92$; the distinct pooled within-model figure is $\rho = +0.777$ at
$n=130$, and we never attribute the cross-model range to a single model. $\lambda$ correlates with
steering effectiveness at $\rho = -0.706$ pooled, reaching $-0.990$ on one family, over 26 layers and 5
families: high $\lambda$ meaning fragile, meaning steers poorly. $A_{\mathrm{lin}}$ is a training-free
top-1 argmax accuracy of the logit lens with no probe fitted, and we reproduce it exactly, applied
through the model's own final norm with a frozen refusal/compliance token family, since refusal-steering
has no native next token.

\paragraph{Altitude.} We never write, at this paper's own altitude, that a cheap forward-pass diagnostic
predicts steering effectiveness: that sentence is owned at this citation. Our forward-pass measurement
forecasts a \emph{cross-context horizontal shift}, a quantity none of these methods forms. Our predictor
is kept provably distinct from $A_{\mathrm{lin}}$ (a soft expected loading of the \emph{steered}
softmax under a continuous signed loading, against hard argmax-membership of the \emph{unsteered} logit
lens) and the distinction is enforced by a unit test rather than asserted, because refuting a
diagnostic with a relabelled copy of itself would be a circular refutation and would deserve to be
caught.

\paragraph{And the arm we called cheap and decisive returned against us.} The registered stop rule fired:
one of these two published within-context diagnostics recovers $\dec$ on this set at the roster's power
(\S\ref{sec:predictor}). The defensible verb for a transferred within-context feature is \emph{does not
transfer to} $\dec$, never \emph{refuted}: neither incumbent claims a cross-context quantity, so a
failure here falsifies nothing either of them asserts.

\subsection{Steerability as a fitted dose--response slope}
\label{app:rel:braun}

\emph{Owned: the steerability score $S$, a fitted dose--response slope. We concede the kinship up front,
and decline to overstate it as an identity.}

\citet{braun2026steerability} defines a steerability score $S$ as a fitted slope of the dose--response
curve, borrowed from \S\ref{app:rel:tan}, to whom the credit for the primitive belongs. The prior measurement's per-model coupling
$T$ is the same kind of object as $S_{\text{agent}}/S_{\text{chat}}$, and the body says so in the first
paragraph that mentions $T$. It is not the same quantity, and we do not claim it is: $S$ is a
least-squares slope fitted across a multiplier grid, whereas $T$ is a ratio of effect summaries read at
an operating point, so the two agree only where the response is proportional and linear. That is
precisely the condition this paper's dose--response fits reject, which is why the distinction is worth
keeping rather than rounding off. Our estimand is deliberately not a slope
ratio: a slope ratio is dose-dependent and dies against a ceiling, whereas a horizontal displacement is
estimable where a vertical gain ratio is not. Conceding the kinship is what lets the switch be argued as
a repair rather than a rebrand; overstating it as an identity would concede more than is true, and would
sit badly beside our own finding that a ratio read at one operating point does not identify
potency and efficacy.

\paragraph{His geometric covariates, stated as the position actually stands.} The two published
predictors of steerability there are a directional-agreement covariate (the mean cosine between each
per-example activation difference and the resulting steering vector) and the discriminability index
$d'$ of positive and negative activations projected onto the difference-of-means line. Both are
properties of the extraction pool and of the direction; because this design holds the direction fixed
while the context varies, neither is a within-cell explanation of $\dec$, and they can only explain
between-model variation in it. That test is pre-registered rather than asserted: both covariates, in
both deployment frames, on the pinned roster, scored against $\dec$ under the same frozen
leave-one-model-out rule the other arms use. \textbf{At the time of writing that measurement has not
been run, and this paper makes no claim to have conditioned on those covariates.} We note in advance
that the roster supports $n=4$ models and that no rank correlation at $n=4$ can reach conventional
significance (the exact two-sided permutation floor is $0.0833$) so the conditioning is
roster-limited by construction. That is a limit we state, not a result we defer.

\subsection{``Context-dependent steering direction''}
\label{app:rel:svf}

\emph{Owned: the phrase. We concede it by name.} \citet{svf2026context} introduced the phrase
``context-dependent steering direction'', and their Fig.~5 is a per-example slope. We concede the phrase
explicitly and mark the difference in object: theirs is a per-example directional quantity in one
setting, ours an aggregate cross-context \emph{location} of a fixed direction carried into a second
deployment format. We do not reuse their phrase for our estimand.

\subsection{Layerwise-Jacobian and subspace transfer instruments}
\label{app:rel:skif}

\citet{skifstad2026jacobian} own a layerwise-Jacobian instrument and a subspace transfer metric, and
measure within a single deployment format, so the cross-context ratio reported here is identically
$1$ in their setup and there is no two-context object to preempt. We pre-empt the obvious appendix
question directly rather than wait to be asked: the contribution is the cross-context comparison their
instrument is not run across, not a better within-context instrument.

\subsection{Cross-architecture steering transfer}
\label{app:rel:agarwal}

\emph{A complementary axis.} \citet{agarwal2026crossarch} varies the target
\textbf{model} with a translated concept direction over all 20 ordered pairs of five models, so the unit
of analysis is the directed model pair. This paper varies the deployment \textbf{context} with model and
vector held fixed, so the unit is the context pair. The axes are orthogonal. Two boundaries keep the
paragraph from becoming a fight.

\emph{Altitude of claim.} That headline is a \textbf{win rate}: whether an effect occurs, at the best
positive strength per pair: $71.0\,\%$ against $68.0\,\%$ native, $65.7\,\%$ baseline and $67.3\,\%$
universal. Ours is a signed magnitude and direction of a location shift. A win rate cannot be confused
with a displacement, so there is no overlap to concede.

\emph{Statistical honesty, at our own bar.} We cite the cross-model advantage as that paper's own data
reports it: directional but not significant, paired $t(4)=1.85$, $p=0.14$, with the abstract itself
calling the gap non-significant. Presenting it as an established effect would violate the standard we
hold ourselves to, and the honest version \emph{strengthens} the estimand boundary: a binary quantity
its own $N=5$ leaves underpowered is precisely the small-$N$ regime a leave-one-family-out design is
built to survive. That work runs a 9-point strength sweep and plots raw strength--response curves but
fits no dose--response location, which is the accurate sentence and the one we write.

\subsection{The Cylindrical Representation Hypothesis}
\label{app:rel:crh}

\emph{Not a novelty threat (single context, per-sample unit, never forms a cross-context object) but
a validity threat, and handled, an asset.} \citet{gao2026cylindrical} hold that steering sensitivity is
controlled sample-specifically, so the per-sample quantity we aggregate has high variance by
construction. Two facts from the full text turn that into support for the design.

First, \textbf{the variance is real and geometrically unpredictable at the sample level}. That paper
reports no quantitative variance statistic, and its own third implication test finds difference-vector
similarity does \emph{not} predict steering behaviour (Pearson $-0.034$, $p>0.05$; a robustness check on
Qwen2.5-14B gives $0.019$, $p=0.297$). So the per-sample susceptibility variance the design must be
powered against is not importable from that paper, and we estimate it empirically from our own
chat-only pilots; the family-level contrast is powered against \emph{that}, never against cell count.

Second, it supplies \textbf{a published mechanism for why the aggregate is the tractable unit}. If
per-sample geometry does not predict per-sample outcome even within one context, a per-example predictor
is in trouble by that paper's own result, and the stable object is an aggregate, which is what $\dec$
is. We cite it as corroborating mechanism for the predicted failure of the per-example baseline, and
carefully: it shows per-example geometry does not predict per-example outcome within one context, while
the arms concern per-example predictors of a cross-context quantity. This is evidence about mechanism and does not constitute proof.

\subsection{The predecessor measurement}
\label{app:rel:prior}

The systematic chat-to-agent transfer measurement this paper reinterprets \citeprior{} established that
the chat-extracted direction survives into ReAct deployment at near-full residual strength while the
behavioural coupling $T$ is reset per model, and (to its credit) publicly refuted its own
pre-registered baseline-shift predictor out of sample. We cite that refutation as prior art and as a
credibility asset. This paper is not the extended version of it: it changes the estimand from a
dose-dependent gain ratio to a horizontal location shift, motivated directly by the theory of
\S\ref{app:rel:taim}, establishes that the ratio does not identify potency and efficacy on grids built to support a
midpoint,
and reports a registered forecast of the location shift that was refuted out of sample. Its published
numbers are reinterpreted under the new estimand; its claim is not extended.

\section{Methods, in full}
\label{app:methods}

The body states the estimand and the scope conditions in one section. This is the complete protocol.
Every procedure below is fixed in a dated pre-registration committed before the run it governs; the
register in \S\ref{app:register} names which one, and is machine-scored against this text rather than
trusted.

\subsection{The estimand}
\label{app:m:estimand}

\paragraph{Why not the gain ratio.} The prior measurement on this question reports a per-model
behavioural coupling $T = \Delta_{\text{agent}}/\Delta_{\text{chat}}$: the ratio of steering-induced
behaviour change in the ReAct-agent context to that in the chat context, read at a matched operating
point. $T$ is not wrong. It is a well-defined functional of two curves and the prior paper stated the
operating point at which it was read. The claim made against it is narrower and it is an
\textbf{identification} claim: \emph{a scalar read at one operating point does not determine the pair
(potency shift, coherent efficacy)}. We establish it by reconstruction rather than by argument. The
prior estimator is re-implemented in both of its published forms: per-dose effect, single-dose ratio,
AUC-over-doses ratio, and the sub-saturation filter at margin $0.05$: copied from its own source, and
re-run on grids swept to their plateaus. Every per-dose effect recomputed reproduces the committed fit's
own record exactly (maximum absolute difference $0.0$, 7 of 7 cells), so what the reconstruction shows
is a property of the estimator and not of a re-implementation.

Two of \emph{our own} registered predictions were refuted in that section and both constrain what the
paper may say downstream. On grids built to the dose rule, $T$ is computable in every cell (34 of 48
cell$\times$dose pairs are defined and no cell yields none) so the prior paper's three families that
``yield no coupling estimate'' are a \emph{grid} property and not a model property. And the
identification level set contains zero in only 2 of 3 scorable cells against a registered threshold of 3.

\paragraph{The estimand and its poolable companion.} $\dec = m_{\text{agent}} - m_{\text{chat}}$, where
$m$ is the fitted 4PL midpoint of the registered combined-binary response in each deployment context, in
commanded-coefficient units. Its scale-free companion is $\Lam = \log m_{\text{agent}} - \log
m_{\text{chat}}$. Theory motivates each property of that choice and we say which result
motivates which, and where the step from their result to our design is ours rather than theirs:
Thm 3.6's reparametrization acts on a curve location, which is why we estimate one; Thm 3.8's
vanishing first-order term is proved for the training-set cross-entropy, and it is
\emph{our} reading, not theirs, that a sign-symmetric first-derivative susceptibility will vanish
likewise, so we make the readouts signed and directional, which a midpoint difference is; Prop 4.1 makes any cross-context object lose identifiability at large dose, and
a midpoint (unlike a top-dose ratio) can be checked against that ceiling, and is.

\textbf{Two dose axes are not commensurable across model families}, which is why $\dec$ is reported per
cell and $\Lam$ carries the cross-family panel: $\Lam$ is invariant to a rescaling of the dose axis where
$\dec$ scales with it. $\Lam$ is \textbf{not new to this project and is not claimed as new}: the same log
ratio was computed weeks earlier, behind a stricter gate, by a different instrument, and the later
instrument reproduces it to the printed digit in 5 of 5 cells. The independent reproduction is reported
as a reproduction.

\paragraph{What the estimand is not claimed to be.} Two registered tests bound the claim from opposite
sides and both are reported whichever way they come out. A shape-invariance test, registered before the
fit, compares a horizontal reparametrization against a vertical rescaling \emph{at equal model
complexity} and reports $\Delta$AIC intervals per cell. And a translation test registers
$\Delta_q = \mathrm{dEC}_{0.75}-\mathrm{dEC}_{0.25}$, which is zero for a pure translation: the paper is
bound by its outcome, and because that outcome rejects, no sentence in this paper may call the two curves
translates of one another. The same registration carries the upper-plateau
comparison. That comparison establishes only that the fitted asymptotes agree; it does not establish
equal achievable effect, and our own therapeutic-window audit says why it cannot: on a
refusal-bypass cell the asymptote is the floor the refusal rate is driven to, so the contrast is
blind to how much refusal there was to remove.

\subsection{Contexts, the control spine, and the metric}
\label{app:m:spine}

The contrast is between two deployment contexts related by a \textbf{matched-information template
ladder}: \textbf{C0}, the bare chat turn, and \textbf{C3}, the same task wrapped in a ReAct scaffold with
deterministic dummy tools of fixed names, signatures and return values.

\paragraph{The match is exact on the task and asymmetric on one span, and we state both.} Audited on the
committed model-facing prompts across all seven probe cells (six distinct models) $\times$ 200 items
$\times$ both contexts, the task instruction is \textbf{byte-identical}, occurs exactly once in each
context, and the rendered prompts are byte-equal outside the system span: 1400 of 1400 item checks,
each rebuilt from its pinned tokenizer revision. The C0$\to$C3 token delta is a per-cell constant
identical for every item ($+144$ tokens on six cells, $+158$ on the seventh), so the wrapper is exactly
additive. What is \emph{not} matched is one span: C0 carries a compliance-valenced directive that C3 does
not carry at all, so the contrast bundles the ReAct frame with the removal of a compliance instruction.
That non-frame component is the first rung of the predecessor's frame ladder, whose marginal CI spans
zero against a format-wrapper rung whose CI excludes it; we report it as an upper \textbf{bound} and draw no licence from it, because it was measured on $T$ at a single dose on one family and arm. Prompt \emph{length}
cannot be matched (the agent prompt is $3.3$--$5.1\times$ longer by construction) which is why the
length control acts on the \emph{output}.

\paragraph{The spine as executed, with its errata carried rather than buried.} Appendix~\ref{app:errata}
states the four record-level errata; here is where each one bites.

\begin{itemize}\itemsep2pt
\item \textbf{A: matched-norm random directions, $\ge 5$ per cell}, swept at every dose in both
contexts on the same item identifiers. They enter as the per-dose reported diagnostic
$\Delta(\text{real})-\max|\Delta(\text{random})|$ and as the $\alpha^{*}$ separability screen; the fitted
response is \textbf{not} the random-corrected one, and applying the correction as a sensitivity leaves
sign, CI-exclusion and status invariant in 5 of 8 cells. Control A is a per-dose diagnostic and a gate,
never a correction of the fitted response.
\item \textbf{F: KV recomputation at every turn}, verified by execution against the pinned generator
source with five predicates each shown able to fail. It guarantees turn-uniform, position-uniform
\emph{re-application}; it is \emph{not} what excludes the prompt--activation duality account, because our
steering is turn-constant and a correct cache would be numerically equivalent. \S\ref{app:m:defl}
registers the missing half as a measurement. \textbf{The guarantee is bounded by how often a rollout
actually reaches the turns it covers}, and the registered rule is that it may not be quoted without that
bound: at the agent midpoint the proven five-turn binding lower bound is $0.729$ on Llama-3.1-8B and
$0.639$ on \texttt{OLMo-2-7B:bypass}, but $0.040$ on the weakest $\alpha^{*}$-GREEN cell. The roster
weakest is \texttt{OLMo-2-7B:induce} at $0.000$, and exact turn depth is not recorded.
\item \textbf{M: the combined binary} (refused \emph{and} coherent), never judge-raw. The inherited
name \emph{length-controlled} does not hold on the property it asserts: the correction the judge applies
is length-correlated as registered but \textbf{one-directional}, so it strengthens rather than removes the
metric's own length association. A second, estimand-specific limit is measured rather than argued: a
context-differential one-directional correction displaces the two midpoints by different amounts:
\emph{the same signature as $\dec$ itself}. The correction is therefore transported to act identically in both
contexts and the frozen fitter re-run on the transported records, gated by two identity checks. Sign is
stable 5 of 5 and CI exclusion 4 of 5, and the one cell that moves moves \emph{toward} exclusion. No
canonical number moves, and the registration forbids reporting a transported value as the paper's $\dec$
under every branch.
\item \textbf{G: the chat gate}, a swing of at least 25 points per family per arm. The clause
\emph{``and beats matched random''} \textbf{never ran}: admission gates on the bootstrap CI lower bound of
the chat swing alone, proven by mutation, and the generator's own two-clause field has zero consumers.
Applying the missing clause now removes exactly one cell, which one is metric-dependent, and \textbf{no
sign-carrier is removed under either rule}. A gate failure is a \textbf{drop after diagnosis}, never a
re-run at a bigger coefficient: the inherited lesson is that a baseline refusal of $0.200$ caps the
achievable swing below the gate, and no coefficient clears a floor.
\item \textbf{N and BS: greedy decoding, 200 items per cell with identical item identifiers, and a
$B=10\,000$ item-paired bootstrap.} Greedy is verified on all $137\,600$ canonical rollout records. The
\emph{fitted} set is smaller and dose-dependent, because coherence-gated rollouts are dropped before the
fit: the resampled anchor is 195--200 and the per-dose paired count falls as low as 0. And the pairing,
while genuinely implemented, is \textbf{inert for this estimand}: $\dec$ is a difference of fitted
\emph{locations}, which does not inherit the cancellation a difference of means at a matched dose does.
\end{itemize}

\textbf{Arm selection is pre-committed per family} from precheck baselines, choosing the arm (induce
(raising refusal) or bypass (lowering it)) with dynamic range in \emph{both} contexts. Where a family
has range on one arm only, that arm is frozen in advance. \textbf{Statistical honesty:} a sign claim
requires the bootstrap CI to exclude zero; we never claim significance where an interval spans the null,
and a null is reported as a null.

\subsection{Dose grids, the validity ceiling, and admission}
\label{app:m:grids}

Fitting a midpoint to a narrow grid is the single most expensive way to be wrong here, and the inherited
grids demonstrate it: the predecessor's coupling grids cluster just under saturation, two points about
ten percent apart, and cannot support a midpoint fit at all.

\textbf{The dose rule: at least five doses spanning at least $4\times$, bracketing the fitted midpoint,
in \emph{both} contexts, or the cell is not fitted.} No full grid is committed blind: a cheap
chat-only pilot locates the transition first, and the paid grid is built from a frozen width prior
around it.

\paragraph{The opposite-sign pair, and the comparison set it was drawn from.} The overlapping-gain pair reported in \S\ref{sec:nonid} is an \emph{existence} result and we give its denominator: the registered scan compares every negative cell$\times$dose against every positive one, $80$ pairs in all, of which $9$ have overlapping gain intervals and $5$ have both intervals tight. Every one of the $9$ uses the sole positive cell, but their negative sides span three distinct cells (OLMo-2-7B bypass, Qwen2.5-7B bypass, Qwen2.5-7B induce), so the search ranges over negative cells as well as over doses. The claim the pair supports is that a gain can coincide across opposite potency shifts, not that it commonly does.

\textbf{What that rule costs, censused over the predecessor's entire raw corpus rather than argued
from one configuration per family.} The paragraph above was originally read off the coupling
configurations alone, which is a sample of the corpus and not the corpus. We have since streamed all
$79$ of the predecessor's rollout files end to end --- $349\,843$ records --- and scored every grid in
them against the rule: $0$ unparseable lines, $0$ records whose deployment context the declared map
could not classify, and $0$ disagreements between the dose named in a run's setting label and the
injected coefficient's magnitude. The corpus holds $73$ (model, arm, layer) cells, of which $25$ carry
a dose grid in \emph{both} deployment contexts. \textbf{Exactly one of the $25$ meets the rule in both,
and it is the cell whose midpoints the predecessor already published} (Qwen2.5-7B, induce, $L16$),
passing inside that configuration's own file --- the only file in the corpus that passes alone. Of the
$22$ two-context pairs outside the three configurations that already carry a published midpoint,
\textbf{none} does. The instrument recovers the one known-good cell and nothing else, on thresholds
fixed before the corpus was read. This is the answer to \emph{why pay for new sweeps instead of
refitting what is already published}: refitting was not an alternative that existed. The roster is
small because the corpus is, not because the search stopped.

\textbf{The honest qualifier is a single near miss, and it does not become a free cell.}
Llama-3.1-8B on the \emph{bypass} arm carries $5$ chat doses spanning $8\times$ and $4$ agent doses
spanning $4\times$: it fails on the counting clause, in the agent context only, by one dose. That is
the pair we would most want, because this paper carries Llama on the induce arm alone, where it is the
roster's only positive cell, so a bypass cell would be a second arm on the positive anchor. It is not
one. The response rate is $0.72$ at dose $4$ and $0.000$ at $8$, $12$, $16$ and $32$ in \emph{both}
contexts, so the whole transition lies inside $(4,8)$ and no swept dose lies inside it; a fifth agent
dose anywhere outside that interval satisfies the clause and carries no information about the
midpoint. The pair also exists only as a union of seven files whose item universes range from $40$ to
$200$, and no single file passes. We report two further readings rather than choosing the flattering
one. The rule as stated is \textbf{stricter than two of the three predecessor configurations that do
publish a midpoint} ($3$ doses over $4\times$; $5$ doses over $2\times$), which makes a null under it
easier to obtain than under the exemplars' own shape; relaxing it to that weakest endorsed shape
($\ge 3$ doses spanning $\ge 4\times$) admits exactly this near miss and no other pair. Both readings
are descriptive: neither gates anything here, and neither would have licensed a fit.

\textbf{The $\alpha^{*}$ validity ceiling} operationalises Prop 4.1. A cell is admitted only if both
fitted midpoints sit below the binding $\alpha^{*}$. That ceiling is
behavioural, not logit-level, and it is computed with the two contexts \emph{merged}. Writing
$\Delta S(\alpha)$ for the real direction's cross-context success-rate separation minus the median
separation of matched-norm random directions, $\alpha^{*}$ is the first dose \emph{above the peak
of} $\Delta S$ at which $\Delta S \le 0$. The high-side anchor is load-bearing rather than
cosmetic: two ascending curves with different midpoints also separate little at low dose, and that
ramp is not the merged limit this ceiling exists to catch. A within-context saturation candidate is
computed and reported and does \emph{not} bind. \textbf{Three cells have no located ceiling}: for Gemma-3-12B induce, Qwen2.5-3B induce and Qwen2.5-7B induce, $\Delta S$ never returns to $\le 0$ above its peak anywhere on the swept grid, so $\alpha^{*}$ is undefined for them and every swept dose is in-regime. Their GREEN therefore records \emph{no ceiling reached on this grid}, a weaker statement than the GREEN of a cell whose midpoints were checked against a located ceiling, and one bounded by the grid we swept rather than by the model. We name the ceiling this way rather than as a
prompt-dependence regime, because the estimator reads rates and never reads logits.
That screen sets the in-regime dose set and each cell's
verdict. A within-context diagnostic is computed per context and reported
alongside it. That diagnostic fires at the lowest dose where \emph{either} the signed real effect
stops exceeding the matched-norm random band \emph{or} the fitted upper asymptote is reached, and
the second branch is not decorative: \texttt{OLMo-2-7B}'s chat leg has a positive real-minus-random
margin at every dose and its diagnostic still fires, at $30$, on saturation alone. It does not bind.
Llama is $\alpha^{*}$-GREEN at $\alpha^{*}=9$ with that diagnostic reading $1$. $\alpha^{*}$-RED cells are
\textbf{printed and excluded from every law claim}, and from the heterogeneity specification this
paper reports (model unit, $\alpha^{*}$-GREEN only). They are law-excluded rather than merely
flagged, and they bind no frozen hyper-parameter. \textbf{Registered aggregates that do contain them, which we do
not claim otherwise}: the fitter's \texttt{power\_crh} block keeps one
representative per \emph{fitter label} (the cell with the smallest standard error). There are seven
such labels, which is neither the eight cells nor the six distinct base models: Qwen2.5-7B's two arms
carry different labels and are never collapsed, and the artifact records this as
\texttt{k\_is\_neither\_unit}. Both $\alpha^{*}$-RED representatives survive that reduction, and
the $\dec$ forest figure's title reports the block, labelled on its face; and the canonical verdict block
counts every fitted cell without an $\alpha^{*}$ filter, so RED \texttt{Starling-7B:bypass} raises
\texttt{n\_families\_fitted}, RED \texttt{OLMo-2-7B:induce} enters \texttt{pos\_sign\_cells}, and
the two-sidedness and phase flags are unfiltered. We read no law from either. The GREEN-only two-sidedness test is
reported separately and passes on its own, which is the version the law rests on. This is a weaker
position than the registration's blanket exclusion and is stated rather than papered over. The exception is
diagnostic and labelled as such: the $\dec$ forest figure's title reports the heterogeneity block
\emph{as the fitter emitted it}, which includes them, and says so on its face so it cannot be read as the registered
figure.

\textbf{Is the fitted shift resolved by the grid, and does its sign need the 4PL?} A registered
resolution check requires the displacement to be larger than the grid can hide, and a fit-free reading of
the same sign means the headline does not rest on a parametric form. The estimator fits the full swept
grid and reports $\alpha^{*}$ post hoc; it does \emph{not} apply the stricter dose-exclusion sentence that
two of our own pre-registrations also register. Re-run under that rule (same estimator, seed, $B$ and
items) the roster's only positive cell keeps its sign and loses CI exclusion. The registered primary
does not move, because the full-grid rule was in force before any paid data existed, but the number that
would change a reader's mind is reported in the results and not only in the limitations.

\textbf{The random arm's item universe is itself a registered choice}, because the separability screen
reads $\max|\Delta(\text{random})|$ and a narrower universe makes the screen easier to pass. The
sensitivity is reported beside the $\alpha^{*}$ analysis rather than folded into it.

\textbf{The grid that was swept is not the grid that was fitted, so the rule is re-checked on the
realized support.} Each context's dose vector is derived downstream from the rollouts that survived
scoring, and a rollout is scored only if its metric field is in $\{0,1\}$ \emph{and} it was coherent.
A dose at which every rollout in one context has collapsed therefore produces no key and leaves that
context's grid with nothing recording that it was swept, which the launch predicate cannot see,
because it is a property of the realized data. The weight has the same shape: the weighted 4PL floors
a bootstrap sd below a minimum so that a \emph{saturated} dose cannot take infinite weight, and the
same line floors a \emph{non-finite} sd, so a dose whose uncertainty could not be computed is given
the smallest uncertainty in the vector. We therefore re-score the realized support of every cell in
the canonical fit on four rails, over law-included cells only: \textbf{(D1)} both contexts are fitted
on the same dose set; \textbf{(D2)} every fitted dose has a finite bootstrap sd; \textbf{(D3)} no
swept dose is removed from a fitted grid by the scoring filter; \textbf{(D4)} the dose rule above
holds on the grid that was fitted rather than the grid that was planned. All four pass on all six
law-included cells and all twelve legs, at a minimum scored $n$ of $26$ per leg.

The one cell at which either condition fires is \texttt{OLMo-2-7B:induce}, which is
$\alpha^{*}$-RED and already excluded from every law claim, and both fire there: at dose $43$ the
chat context has $0$ of $200$ rollouts coherent and the dose leaves its grid, the agent context keeps
$2$ and retains it, and that retained point's undefined bootstrap sd is floored to the minimum. The
cell is discussed as an artifact in the main text; what these rails add is the two-directional
statement that the mechanism reaches no cell the law uses.

\subsection{Estimation, inference, and the analytic multiverse}
\label{app:m:inference}

\paragraph{The fit.} A four-parameter logistic in each context, with $m$ its midpoint. Bootstrap
$B=10\,000$, item-paired, seed fixed in advance. The per-dose bootstrap standard deviation weights the
fit, which is why $B$ is not a cost dial: lowering it moves the fitted location rather than only widening
the interval.

\paragraph{Is the interval calibrated?}
\label{app:calib}
A registered simulation calibrates the reported $\dec$ interval
against known ground truth on the realized grid geometry, with the repair (recomputing the per-dose
weights \emph{within} each resample) registered \textbf{conditionally} before any replicate was drawn,
and applied to all law-eligible cells or to none. Point estimates move by less than $5\times10^{-4}$ and
no cell's CI-exclusion status changes. Prereg \S6.1 requires \emph{both} intervals published, the committed one dated rather
than deleted, so both are printed here. Writing committed $\to$ repaired:
\texttt{Llama-3.1-8B:induce} $+1.013$, $[+0.777,+1.273] \to [+0.760,+1.295]$;
\texttt{OLMo-2-7B:bypass} $-5.497$, $[-7.851,-3.175] \to [-7.687,-3.045]$;
\texttt{Qwen2.5-3B:induce} $-0.237$, $[-1.303,+0.818] \to [-1.358,+0.818]$;
\texttt{Qwen2.5-7B:bypass} $-10.855$, $[-15.814,-4.614] \to [-15.790,-4.253]$; and
\texttt{Qwen2.5-7B:induce} $-12.368$, $[-13.726,-10.876] \to [-13.915,-11.017]$. Four cells clear
zero before and after, the one that never did still does not, and Table~\ref{tab:core} prints the
committed interval because that is the one the registration dates. Three cells carry no repaired
interval, for two different reasons,
and the distinction is stated rather than smoothed: two are $\alpha^{*}$-RED and outside the repair's
frozen roster, while one is law-eligible but \emph{post-dates} that roster, which was frozen before the
seventh family existed. Re-running the repair to admit it would re-open a roster frozen before any
replicate was drawn, so the gap is named rather than closed by amendment.

\paragraph{Power, at the unit the paper reports.} Because steering sensitivity is controlled
sample-specifically (\S\ref{app:rel:crh}), the family-level contrast is powered against \textbf{per-sample
susceptibility variance} rather than against cell count, and the per-sample variance is reported here rather
than only asserted, with its mean beside it as the registration asks: across the sixteen
cell-context estimates the variance spans $0.130$ to $0.278$ and the mean spans $-0.285$ to
$+0.809$. The scorer previously stored only the variance, so an earlier version of this paper
reported the range and stated that no mean was available; the scorer now records both, and the
refit that produced them is bit-identical to the committed artifact in every other field. \textbf{The unit of analysis is the model, never the cell} (two arms of one model are one
model) and every roster count in the paper is stated at that unit.

\paragraph{The selection bound, signed and measured.} Rather than hedge about the roster's admission
geometry we measure it. All three agent-side measurability gates are monotone in absolute dose, so the
admission window reaches less far above zero than below it, and the design under-observes exactly the sign
that is scarce. The observed positive rate is therefore reported as \textbf{biased downward by the
admission geometry} (not as a bound on a population rate, which five cells drawn under no sampling
frame cannot support) with the share attributable to our own width prior separated out ($63.0\,\%$ of the pooled total).

\paragraph{The analytic multiverse.} A registered specification curve varies response level, link
function, fit window and metric, and reports the \textbf{sign agreement rate} across specifications. The
defensible claim is about the \emph{sign}: $\mathrm{dEC}_{0.50}$ varies by up to $2.99\times$ across links
and the AIC-preferred link is our registered 4PL in only 1 of 5 cells: a prediction our own
pre-registration made \emph{against} our registered choice, and it held.

\paragraph{Coherence is fitted as an outcome and never used to exclude a cell.} The coherent margin
$M = m_{\text{coh}} - m$ is registered as a second dose--response dimension: where the behavioural
midpoint sits relative to the dose at which coherence half-collapses. This is what makes a fitted midpoint
readable as behaviour rather than as decoding failure, and it is the instrument that identifies the one
cell this paper quarantines.

\subsection{Roster, and how the seventh family was chosen}
\label{app:m:roster}

The backbone roster is inherited (continuity is what makes this a reinterpretation of published numbers
rather than a cold start) and every family passes the chat gate on a pre-committed arm or is dropped
before it consumes a verdict.

\textbf{The seventh family was selected under a registered protocol that exists to defeat one specific
objection}: choosing a family \emph{because} our forecast scores it well and then validating the forecast
on it. The registration therefore fixes, before the run: the predictions of \textbf{all three} forecasting
arms for the chosen family, so the new datapoint is a fair three-way test and not a showcase; selection on
\textbf{predictor-definedness} as well as on predicted sign, so a family predicted to sit at its own noise
floor is not bought; and the \textbf{full candidate ranking}, with each arm's forecast for each candidate
and the reason the winner won, so the selection is auditable. A candidate that failed definedness was
dropped under that rule rather than run in hope.

\subsection{Deflationary accounts, measured before generation}
\label{app:m:defl}

Four mechanisms could produce a context-dependent horizontal displacement with no reparametrization at
all. Three are measured and rejected; the fourth is excluded by construction, and the difference in
logical status is stated in the first sentence of its subsection rather than left to the reader. Each of
the three is measurable from forward passes alone, before any generation, and each was registered before
the quantity it measures existed anywhere in this project.

\paragraph{Residual-norm rescaling.} \emph{The account:} the agent prompt has a larger residual norm, so a
fixed $\alpha$ is a smaller relative perturbation, so the midpoint moves: a pure scale effect. \emph{The
measurement:} the norm ratio $\rho = \|r\|_{\text{agent}}/\|r\|_{\text{chat}}$ is estimated per cell with a
CI and its \emph{sign pattern} compared against $\Lam$'s. The registered falsification is structural rather
than quantitative: a scale account predicts \textbf{one sign for every model}, so a two-sided $\Lam$
refutes it whatever the magnitudes.

\paragraph{Baseline alignment.} \emph{The account, and it is the sharper one:} the norm law is the
$c_0 = 0$ special case of the general drive
$\cos\theta_v(g) = (c_0+g)/\sqrt{1+2c_0g+g^2}$ with $g = \alpha/\|r_L\|$. If the unsteered residual is
already partly aligned with the steering direction and that alignment differs by context, the composed
scale-plus-translation account generates a displacement with no reparametrization. \emph{The measurement},
committed before any cosine between a residual and a steering direction had been computed once in this
project, and CPU-only: four registered questions (is $c_0$ nonzero, does it differ by context, does the
\emph{composed law} reproduce the measured $\Lam$, and does the affine form in the literature fit) each
with its rejection threshold registered in advance. The composed-law question carries a \emph{count}
threshold, and its verdict is quoted as ``$k$ of $n$ scorable cells reject, against a rejection threshold
registered in advance'', never as a bare ratio implying unanimity.

\emph{Scope, shipped in the artifact's own field.} Every $c_0$ above is read at the \textbf{final prompt
token}, while the deployment operator adds $\alpha\cdot u$ at \textbf{every} position; a prompt-block
readout also understates generation-time variance. A registered completion removes the first half of that
limit by measuring $c_0$ over the whole steered position set, block-partitioned into the instruction span
the two contexts share verbatim and the scaffold span only C3 has. That partition makes the
instruction-block contrast a context effect on a \emph{fixed token set}: the matched-information control
applied to a geometric readout. A free preflight established that all six roster cells partition cleanly
and that the instruction block is token-identical across contexts in every one.

\paragraph{Dose transduction.} \emph{The account:} $\dec$ is a difference of midpoints in
\textbf{commanded-coefficient} units, and there is no reason a commanded unit buys the same residual
displacement in a long ReAct prompt as in a short chat prompt. \emph{The measurement:} the delivered
displacement per commanded unit is measured per context with an interval, the objection is scored on its
own terms, and the location is then \textbf{re-expressed in chat-delivered units} by the algebraic
correction $R\,m_{3}-m_{0}$. No curve is re-fitted, and we do not claim one is. The registered reading
requires all three: whether the axes are commensurable, how much of $\dec$ delivery explains and
\emph{with what sign}, and whether sign and CI-exclusion survive the correction.
\textbf{Scope, binding on every quotation of this result}: $n=3$ cells over two models, one behaviour
family, one dose per cell; $R$ comes from a $120$-item projection replay while $m_{0}$, $m_{3}$ and
$\dec$ come from independent behavioural sweeps, so it is not a paired comparison; and $m_{0}$, $m_{3}$
are normal approximations.

\paragraph{KV-cache reuse: excluded by construction, and then registered as a measurement.} This account
has a different logical status from the three above and the paper says so in its first sentence: those are
mechanisms we measured and rejected, this one is a mechanism our design never admits.
\citet{kang2026duality} attribute multi-turn coherence failure to KV-cache contamination: each generated
token writes steered states into the cache, so later tokens attend to an increasing set of perturbed
states. If a ReAct rollout reused its cache across turns, the agent response could differ for a reason
unrelated to a dose--response reparametrization.

\emph{The design:} every agent-context rollout behind every $\dec$ in this paper runs under \textbf{mandatory
full re-encode at every turn}, verified structurally in both agent-context paths. In the single-item loop
each turn rebuilds the prompt string and passes a list of \emph{strings}, with no parameter through which a
cache could be threaded, plus a hard assertion naming the control; in the batched path, batching is across
items within a turn and recompute across turns within an item, and the two are orthogonal. The check
\emph{can fail} (adding a cache parameter to either call site flips it) which is what makes it a
verification rather than a restatement.

\emph{Three scope limits bind, and all three are stated:} we have \textbf{not} measured what cache reuse
contributes, only that our result does not depend on it; we do \textbf{not} say that account is wrong, only
that the mechanism is untested there and inapplicable here; and recompute guarantees \emph{re-application},
not duality exclusion, because under turn-constant steering a \emph{correct} cache would be numerically
equivalent. \S\ref{app:notrun} describes the registered arm that converts the first limit into a
measurement.

\subsection{Frame, extraction, and injection-site controls}
\label{app:m:frame}

\emph{The objection, and it is mechanical:} the direction is chat-native, so of course it behaves
differently under a ReAct scaffold: you are measuring how badly a chat-fitted probe is mis-specified for
the agent context and calling the mis-specification a reparametrization.

The \textbf{extraction-frame control} answers it on already-existing data. The predecessor project ran a
pre-registered, clean-room-verified reciprocal-extraction matrix in which the same instructions were also
rendered inside the ReAct frame, and this paper reads that matrix on its own estimand for the first time.
Four questions are registered (sign invariance across frames, magnitude stability against a registered
ratio bound, whether the \emph{agent-native} direction \emph{widens} the displacement, and whether the two
arms agree) with the fourth's unfavourable branch written as unfavourable. That registration discloses
in full that the analyst had already seen a point-estimate preview, so it is a \textbf{pre-specified
analysis of already-public data, not a blind pre-data registration}, and it is described that way wherever
it is cited.

\textbf{The injection layer is the other nuisance parameter of a cross-context estimand that was chosen on
one of the two legs}, and a registered sweep bounds what that choice buys. The two incidents (extraction
frame and injection layer) are named in the limitations as a \textbf{class}, not as two unrelated events.
Four further frame controls are registered and reported in their own companion documents: the predecessor's
five-rung matched-information frame ladder read on the horizontal estimand, the same ladder on the refusal
arm, the frame-state ablation and conditional-transport stages, and the injection-\emph{site} confound.

\subsection{The forecast and the two refutation arms}
\label{app:m:predictor}

\textbf{The forecast} predicts $\dec$ from a pre-generation directional susceptibility measurement,
\textbf{leave-one-family-out, with no per-model fitted parameter}. Because Thm 3.8 zeroes the first-order term of the
training-set cross-entropy, and because we read that as extending to sign-symmetric aggregates
generally, the susceptibility is signed and directional by construction; a symmetric readout would on
that reading be identically zero and is used nowhere in this paper.

\textbf{The two refutation arms run before anything downstream depends on the forecast}, because they are
cheap, forward-pass only, and decisive. A published within-context perturbation sensitivity, and a
published logit-lens accuracy, are each fit per context and each scored as a predictor of $\dec$ on the
same leave-one-family-out protocol, against a naive roster-mean baseline. The published \emph{per-example}
geometry classifier is registered and \textbf{not scored}: it needs per-example success labels this roster
never produced. The registered
stop rule is explicit and it fired: \emph{if a published within-context diagnostic also recovers $\dec$ on
this set, the forecast contributes nothing that is separable at this roster size}. That is reported rather
than routed around. An identification gate registered with the arms bounds when a leave-one-out affine map
on a small $N$ is identified against a collinear competitor at all.

\textbf{The out-of-sample test is the one that settles it.} A signed prediction for a family whose $\dec$
did not yet exist (sign, normalised rank, and both refutation clauses) was committed before that
family's sweep ran. Both clauses failed. We report the refutation as the result, we do not describe the
forecast as a validated predictor, and we take no not-applicable escape that the registered branches did
not authorise. \textbf{No entrant is promoted after the fact}: a five-entrant gauge board registered before
the landing is scored as registered, its own reading is gauge-dependent, and the post-hoc winner is not
promoted.

\subsection{Register-then-run, and what the record can and cannot re-run}
\label{app:m:register}

\textbf{Every paid experiment has a design document and a pre-registration with live/die criteria committed
before launch}, and the corpus of those registrations is the denominator of two instruments. One asks
whether every registered \emph{rule} has a recorded \emph{verdict} and where; \S\ref{app:register} asks
whether every registered \emph{design} has a stated \emph{procedure} and where. Both fail closed: a rule
answered nowhere is a defect rather than a silent pass, and so is a design described nowhere.

\textbf{Guards are run as a suite, with the exceptions named.} Each control's verification lives in an
executable guard, and every guard is required to be able to \textbf{fail}: verified by mutation, not by
review. Until late in the project no session had run them as a suite; the first full-suite run returned four
red where the operating log tracked two, and none of the four named the defect it actually had. The suite
now carries a declared environment per guard and a third status, \emph{not runnable here}, because an absent
environment is neither a pass nor a failure. \textbf{We do not claim a uniformly green suite. We claim a
suite that has been run as a suite, with its exceptions named.}

\textbf{Data availability, measured rather than asserted.} The reproducibility census is re-measured against
the tree a reviewer actually receives rather than against the working tree, which holds tens of gigabytes of
rollout records and activation bundles that the ignore file excludes. Fitted $\dec$ values, the
control-spine audits and the forecasts are computed from data that is \textbf{not} in the code repository.
\textbf{We do not claim that the repository reproduces the results}; we state what it does regenerate and
name what it does not. Run-time provenance is a separate and weaker record than pinning, and is reported as
such: a pinned revision records what was \emph{requested}, and only a capture taken while the machine is
alive records what was \emph{loaded}.

\section{The study register: every registered design, and where its procedure is stated}
\label{app:register}

This table is the denominator's counterpart and it is \textbf{scored rather than taken on trust}. An audit takes the
mechanical set of pre-registration documents released with the paper as its denominator; a registration
missing from this table is a defect, a row naming a section that does not exist (or a section that does
not cite it) is a defect, and a row deferring to a document that has since appeared is a defect the
moment it appears, so the deferral clears itself rather than ageing into cover. Each key below names one
pre-registration document, released with the paper. Dispositions: \textbf{main} (procedure in
\S\ref{app:methods}); \textbf{companion} (procedure in the named companion document, which must cite
its own pre-registration); \textbf{infra.} (operational design with no reviewer-facing protocol);
\textbf{not run} (registered, never run, reported as such, \S\ref{app:notrun}).

The register reports \textbf{47 of 47} with \textbf{0} defects, disposed as 25 stated in the full methods
appendix, 18 in the named companion documents, 2 infrastructure and 2 registered-but-not-run. It was at \textbf{10 of 44} when it was first measured, which
is the number Appendix~\ref{app:apparatus} reports and the reason the instrument exists.

{\footnotesize
\setlength{\tabcolsep}{4pt}
\begin{longtable}{@{}p{0.245\textwidth}p{0.105\textwidth}p{0.195\textwidth}p{0.395\textwidth}@{}}
\caption{The study register: all 47 registered designs, where each is reported, and what it
fixes.}\label{tab:register}\\
\toprule
registration & where & section & what it fixes \\
\midrule
\endfirsthead
\caption[]{The study register \emph{(continued)}.}\\
\toprule
registration & where & section & what it fixes \\
\midrule
\endhead
\texttt{week1\_\allowbreak dose\_\allowbreak grid} & main & \S\ref{app:m:spine} & the two-context sweep, the control spine and the arm-selection rule \\
\texttt{scalar\_\allowbreak gain\_\allowbreak nonidentification} & main & \S\ref{app:m:estimand} & the reconstruction of the prior estimator on our grids \\
\texttt{scale\_\allowbreak commensurability} & main & \S\ref{app:m:estimand} & the identity gate under which the log ratio may be pooled \\
\texttt{potency\_\allowbreak identification} & main & \S\ref{app:m:estimand} & translation test and upper-plateau comparison \\
\texttt{shape\_\allowbreak invariance} & main & \S\ref{app:m:estimand} & horizontal against vertical at equal model complexity \\
\texttt{alpha\_\allowbreak star} & main & \S\ref{app:m:grids} & the Prop 4.1 validity ceiling and the separability screen \\
\texttt{dose\_\allowbreak resolution} & main & \S\ref{app:m:grids} & grid resolution, and the sign without the 4PL \\
\texttt{random\_\allowbreak universe} & main & \S\ref{app:m:grids} & the item universe of the matched-norm random arm \\
\texttt{estimator\_\allowbreak calibration} & main & \S\ref{app:m:inference} & is the reported interval calibrated on the realized geometry \\
\texttt{estimator\_\allowbreak repair} & main & \S\ref{app:m:inference} & the conditional within-resample weight repair \\
\texttt{power\_\allowbreak analysis} & main & \S\ref{app:m:inference} & powered against per-sample variance, at the model unit \\
\texttt{selection\_\allowbreak correction} & main & \S\ref{app:m:inference} & the admission geometry, measured and signed \\
\texttt{spec\_\allowbreak curve} & main & \S\ref{app:m:inference} & the analytic multiverse and the sign-agreement rate \\
\texttt{coherent\_\allowbreak efficacy} & main & \S\ref{app:m:inference} & coherence fitted as an outcome, never used as a filter \\
\texttt{sixth\_\allowbreak family\_\allowbreak selection} & main & \S\ref{app:m:roster} & three-arm registration, definedness screen, full ranking \\
\texttt{norm\_\allowbreak rescaling} & main & \S\ref{app:m:defl} & the residual-norm account and its sign-homogeneity test \\
\texttt{baseline\_\allowbreak alignment} & main & \S\ref{app:m:defl} & the composed drive law and its registered rejection count \\
\texttt{baseline\_\allowbreak alignment\_\allowbreak allposition} & main & \S\ref{app:m:defl} & the all-position completion and the block partition \\
\texttt{dose\_\allowbreak transduction} & main & \S\ref{app:m:defl} & commanded against delivered dose, and the re-expressed law \\
\texttt{cache\_\allowbreak reuse} & main & \S\ref{app:m:defl} & the reuse leg paired against the committed recompute leg \\
\texttt{cache\_\allowbreak reuse\_\allowbreak policy} & main & \S\ref{app:m:defl} & resolving what "cache reuse" means before the pod \\
\texttt{extraction\_\allowbreak frame} & main & \S\ref{app:m:frame} & the reciprocal-extraction matrix read on this estimand \\
\texttt{layer\_\allowbreak sensitivity} & main & \S\ref{app:m:frame} & what the one-leg choice of injection layer buys \\
\texttt{predictor} & main & \S\ref{app:m:predictor} & the signed directional-susceptibility forecast, LOFO \\
\texttt{refutation\_\allowbreak arms} & main & \S\ref{app:m:predictor} & the two published baselines and the registered stop rule \\
\texttt{judge\_\allowbreak transport} & companion & \texttt{judge\_\allowbreak transport\_\allowbreak appendix} & the context-differential judge correction, transported and re-fitted under two identity checks \\
\texttt{positive\_\allowbreak arm\_\allowbreak replication} & companion & \texttt{positive\_\allowbreak arm\_\allowbreak replication\_\allowbreak appendix} & the positive arm re-fitted on a second, independent dose grid from the parent repository \\
\texttt{format\_\allowbreak ladder\_\allowbreak potency} & companion & \texttt{format\_\allowbreak ladder\_\allowbreak potency\_\allowbreak results} & the matched-information format ladder read as a horizontal displacement rather than a gain \\
\texttt{refusal\_\allowbreak ladder} & companion & \texttt{refusal\_\allowbreak ladder\_\allowbreak results} & the same matched-information ladder run on the refusal arm of the roster \\
\texttt{frame\_\allowbreak scaffold} & companion & \texttt{frame\_\allowbreak scaffold\_\allowbreak results} & the parent's five-rung frame ladder, re-read on the horizontal estimand rather than on the gain \\
\texttt{frame\_\allowbreak transport} & companion & \texttt{frame\_\allowbreak transport\_\allowbreak stage0\_\allowbreak results} & frame-state ablation stages and the conditional transport arm that was gated on them \\
\texttt{phase\_\allowbreak dose} & companion & \texttt{phase\_\allowbreak dose\_\allowbreak results} & the injection-site (phase) confound, measured against the horizontal displacement \\
\texttt{protocol\_\allowbreak integrity} & companion & \texttt{coherent\_\allowbreak efficacy\_\allowbreak results} & agent protocol integrity treated as a context-exclusive response dimension of its own \\
\texttt{generated\_\allowbreak block\_\allowbreak readout} & companion & \texttt{generated\_\allowbreak block\_\allowbreak readout\_\allowbreak results} & the steering axis read over the model's own generated tokens rather than over the prompt \\
\texttt{therapeutic\_\allowbreak window} & companion & \texttt{therapeutic\_\allowbreak window\_\allowbreak results} & the deliverable dose-response, and this project's first efficacy-level rather than location estimand \\
\texttt{baseline\_\allowbreak offset} & companion & \texttt{baseline\_\allowbreak offset\_\allowbreak results} & whether the parent paper's own refuted baseline-shift predictor recovers the horizontal estimand \\
\texttt{gemma3\_\allowbreak scaffold\_\allowbreak gate} & companion & \texttt{gemma3\_\allowbreak scaffold\_\allowbreak gate\_\allowbreak results} & the recoverability gate and reader validity, measured on a fourth family before spending \\
\texttt{gemma2\_\allowbreak fold\_\allowbreak control} & companion & \texttt{gemma2\_\allowbreak fold\_\allowbreak control\_\allowbreak results} & breaking the chat-template fold confound with a 2024-era model that folds the system turn \\
\texttt{qwen3\_\allowbreak generation\_\allowbreak control} & companion & \texttt{qwen3\_\allowbreak generation\_\allowbreak control\_\allowbreak results} & the empty cell of the generation-by-format design, and the first test of format capture there \\
\texttt{q2\_\allowbreak internlm\_\allowbreak feasibility} & companion & \texttt{q2\_\allowbreak internlm\_\allowbreak feasibility\_\allowbreak results} & whether the salvage cell that the gain ratio loses is fundable at all before committing to it \\
\texttt{phase\_\allowbreak b\_\allowbreak sharding} & infra. & --- & how the paid sweep is sharded across pods and co-tenants; a scheduling design with no estimand of its own \\
\texttt{react\_\allowbreak batching\_\allowbreak revalidation} & infra. & --- & re-validation of batched ReAct against the already-accepted noise floor; a harness authorisation, not a reported measurement \\
\texttt{scaffold\_\allowbreak intervention} & not run & \S\ref{app:notrun} & the causal scaffold edit, registered in full and never run: its own attainability analysis showed the registered verdict was unreachable at the affordable design, and the manuscript reports it as the experiment we did not run \\
\texttt{braun\_\allowbreak covariates} & not run & \S\ref{app:notrun} & Braun's geometric covariates against the horizontal estimand, registered with both covariates in both contexts and not yet measured; Limitations states the gap rather than the prereg implying a result \\
\texttt{roster\_\allowbreak currency\_\allowbreak rescore} & companion & \texttt{roster\_\allowbreak currency\_\allowbreak rescore\_\allowbreak results} & the deflationary-accounts re-score that admitted the post-freeze $\alpha^{*}$-GREEN cell as a secondary panel; a record-integrity repair with no estimand of its own, reported because it moved one verdict (\texttt{NR-Q6}) and discharged threat 4.37 \\
\texttt{gemma\_\allowbreak ladder} & companion & \texttt{gemma\_\allowbreak ladder\_\allowbreak results} & the parent's Gemma C0/C2/C3 format ladder read on the horizontal estimand at both dose scales, localising the displacement to the format rung on a second family \\
\texttt{parent\_\allowbreak grid\_\allowbreak census} & companion & \texttt{parent\_\allowbreak grid\_\allowbreak census\_\allowbreak results} & the complete dose-grid census of the parent's raw corpus against the fittability rule: whether any unfitted two-context cell exists to refit, answered over all 349,843 records \\
\bottomrule
\end{longtable}
}

\section{Additional figures and tables}
\label{app:figures}

The body carries one figure, because at nine pages a second float costs a paragraph of the identification
argument. The rest of the registered figure set is here. Every panel is rendered from a committed fit
artifact by a driver whose consumed fields were declared in a figure plan frozen before the fit, and every
null render (what the panel would look like if the result had gone the other way) was written into
that plan in advance. Two figures of the frozen set are \textbf{withdrawn rather than nulled}, and
\S\ref{app:fig:withdrawn} says why, because drawing a pre-written null for an experiment that never ran
would be a fabricated measurement.

\begin{figure}[htbp]
\centering
\includegraphics[width=0.92\textwidth]{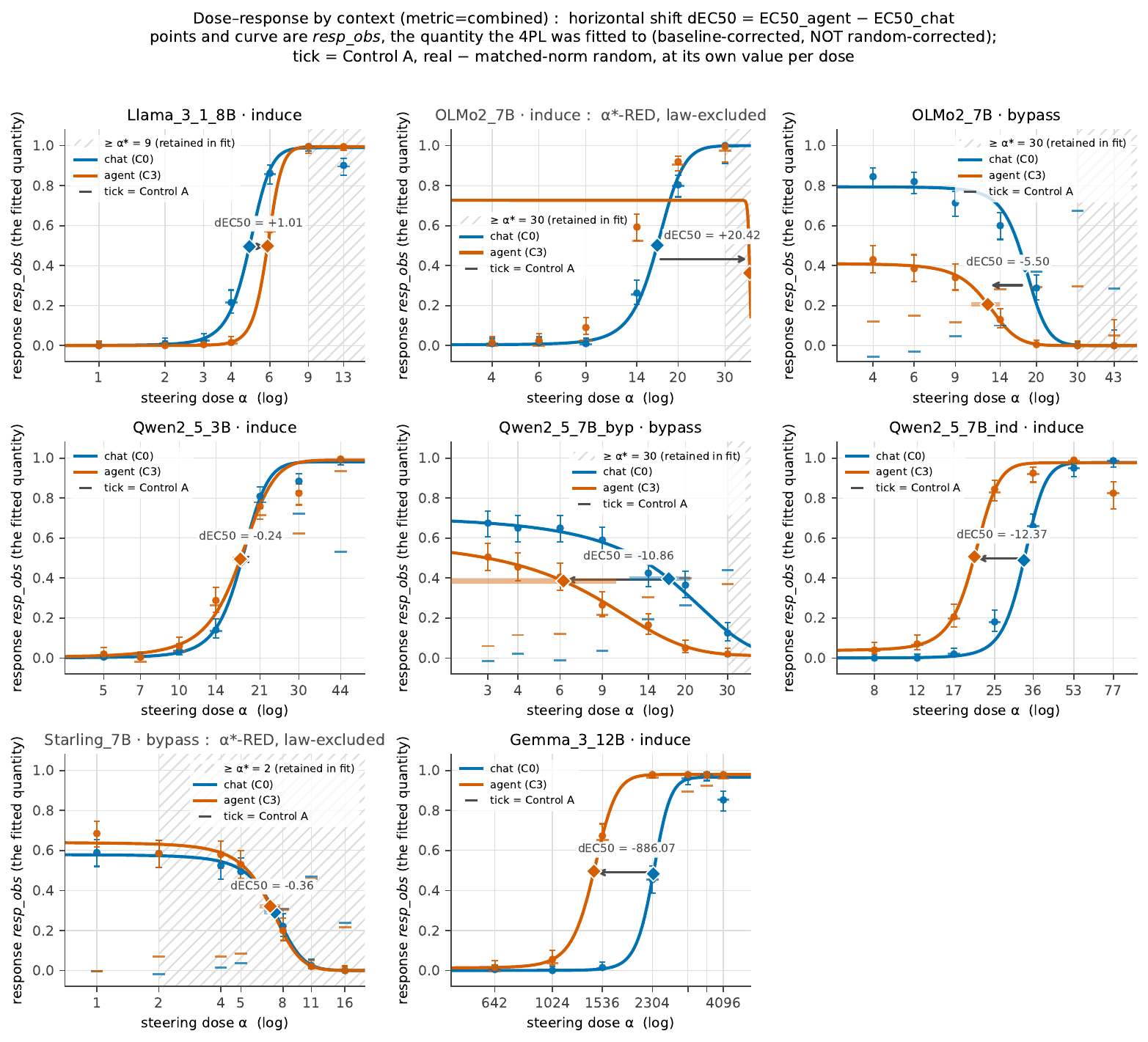}
\caption{\textbf{The law, cell by cell.} One panel per fitted cell; each overlays the chat and agent
observed responses, their fitted 4PL curves, midpoint markers with CIs, and a hatched region for doses at
or above the validity ceiling $\alpha^{*}$. \textbf{Points and curve are the same quantity}:
the response the 4PL was fitted to, which is baseline-corrected and, as errata item 1 records,
\emph{not} random-corrected. Control A is drawn per dose as a separate tick at its own value, so the
matched-norm random comparison is visible without being mistaken for the fitted series. Those doses are \emph{retained} in the fit: the
primary estimator is the full swept grid and $\alpha^{*}$ is reported post hoc (\S\ref{app:m:grids}),
so the hatching marks a scope condition, not an exclusion. The registered null render for this panel is
overlapping curves with coincident midpoint bands: a family with no shift would be reported, not dropped.}
\label{fig:curves}
\end{figure}

\begin{figure}[htbp]
\centering
\includegraphics[width=0.86\textwidth]{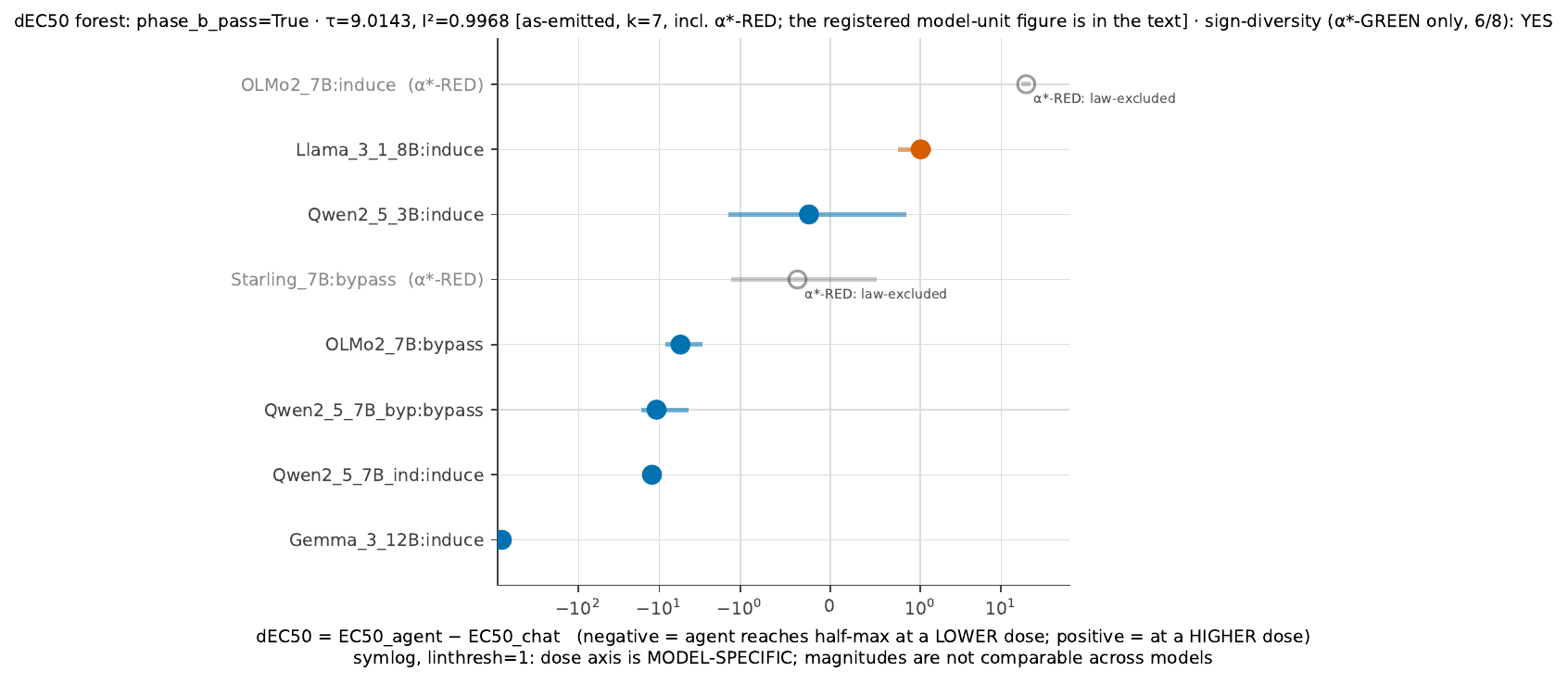}
\caption{\textbf{Forest of $\dec$, and the heterogeneity the aggregate rests on.} One row per fitted cell,
point and item-paired bootstrap CI, ordered by value, with the null line always in frame. The axis is
symmetric-log rather than linear: on a linear axis one cell's model-specific dose units compress every
other row to a point, which would be a presentational choice doing the work of a result. $\alpha^{*}$-RED
cells are drawn in a subordinate style and tagged law-excluded: they stay \emph{in frame}, because
dropping them would be the mirror-image dishonesty, but they cannot be misread as evidence, and the
sign-diversity annotation is computed over GREEN cells only. The legend deliberately does not say
``amplifier'' or ``attenuator'': those are gain words, and $\dec$ is a location.}
\label{fig:forest}
\end{figure}

\begin{figure}[htbp]
\centering
\includegraphics[width=0.95\textwidth]{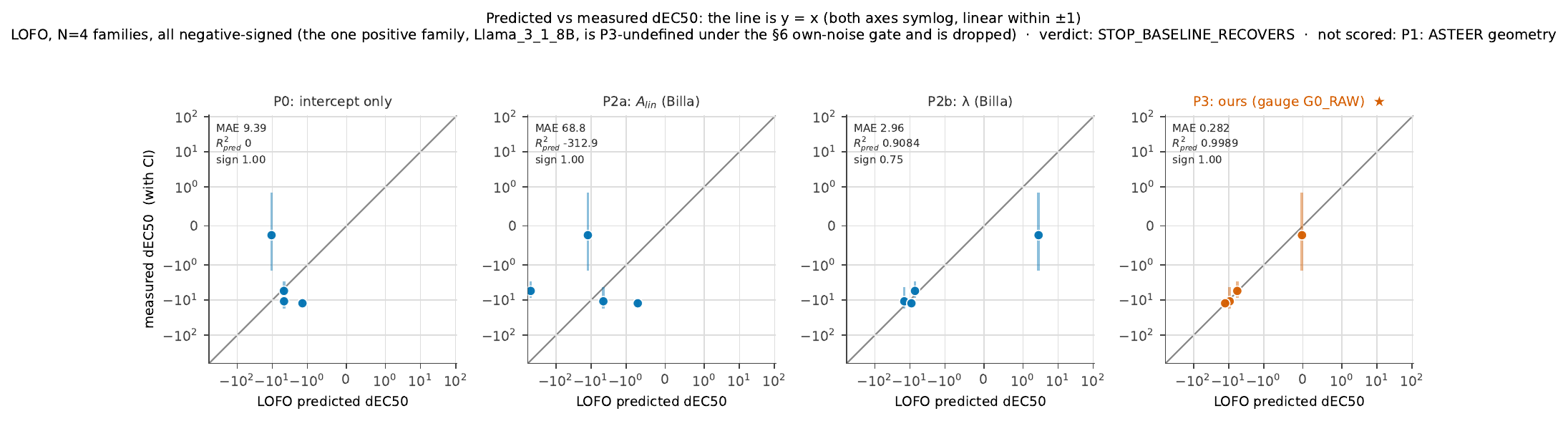}
\caption{\textbf{Leave-one-family-out forecast against measured $\dec$, one facet per predictor.}
\textbf{Our facet's printed MAE and $R^{2}_{\text{pred}}$ are at gauge \texttt{G0\_RAW}
($\kappa=0$) and hold at no other}: every gauge the registered audit calls defensible puts
$R^{2}_{\text{pred}}$ below zero, and the slope's sign moves with the gauge. The
facet set is the four predictors that were actually scored; the per-example geometry baseline was never
scored on this set and the figure says so rather than drawing an empty panel. \textbf{The danger render is
drawn as faithfully as the favourable one}: a baseline hugging the diagonal is the registered stop
outcome, and that is what one facet shows. The $A_{\mathrm{lin}}$ facet's caption labels its published
correlation as the cross-model, within-context figure it is, and states that the diagnostic was built to
predict within-context success rather than to locate a cross-context displacement
(\S\ref{app:rel:fanbilla}).}
\label{fig:predictor}
\end{figure}

\begin{figure}[htbp]
\centering
\includegraphics[width=0.92\textwidth]{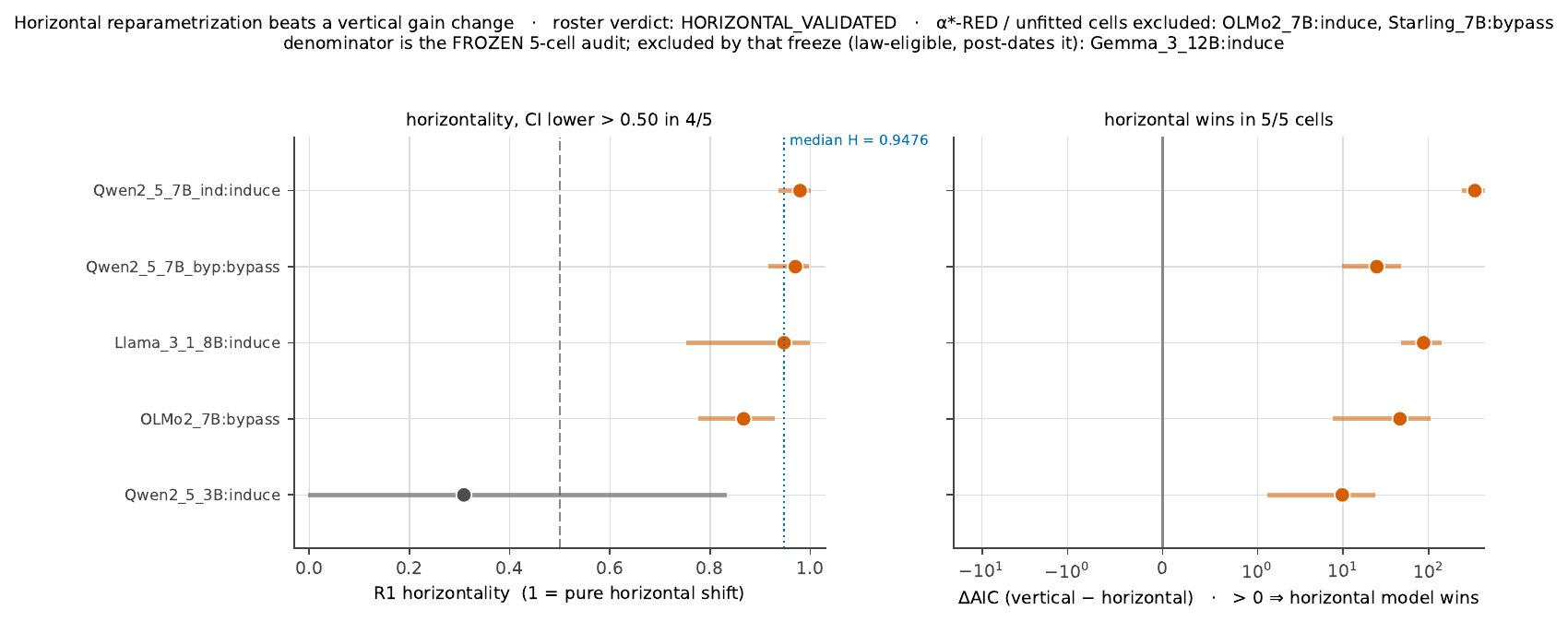}
\caption{\textbf{Horizontal against vertical, at equal model complexity.} \emph{(left)} per cell of the
\textbf{frozen five-cell audit}, the horizontality statistic with its bootstrap CI against the registered
threshold of $0.50$, with the median annotated. \emph{(right)} the same five cells, $\Delta$AIC of the
vertical model minus the horizontal one with its CI, null line always in frame; positive means the
horizontal model wins. \textbf{The denominator is five, not the roster's six $\alpha^{*}$-GREEN cells}:
\texttt{Gemma\_3\_12B:induce} is law-eligible and $\alpha^{*}$-GREEN but landed after this audit was
frozen, so it is named on the figure as excluded-by-freeze rather than left to be inferred from a short
list. Law-excluded and unfitted cells are not drawn (the verdict is defined over GREEN cells) but
their exclusion is likewise named with the cell keys. A CI crossing either reference line would be drawn
crossing it; nothing about the layout presumes the pass.}
\label{fig:horizontality}
\end{figure}

\begin{table}[htbp]
\centering\small
\caption{\textbf{The refutation scorecard.} Leave-one-family-out mean absolute error with its bootstrap CI,
predictive $R^{2}$, and sign accuracy, for the naive intercept-only baseline, the two published
within-context diagnostics, and our forecast. \textbf{The registered PASS condition was that the published
baselines are refuted \emph{and} ours recovers; the realized verdict is that a published baseline
recovers}, so the condition is not met and this is reported as the headline negative. Read sign accuracy
with care: the head-to-head families are all negative-signed, so a constant-sign rule scores 4 of 4 by
construction: which is exactly what the naive roster-mean baseline does at
$R^{2}_{\text{pred}} = 0.000$.}
\label{tab:refutation}
\begin{tabular}{@{}lrlrcl@{}}
\toprule
predictor & MAE (LOFO) & MAE 95\,\% CI & $R^{2}_{\text{pred}}$ & sign & arm verdict \\
\midrule
naive (roster mean)                & 9.39  & [8.06, 10.84]  & 0.000            & 4/4 & --- \\
per-example geometry               & \multicolumn{5}{l}{not scored on this set} \\
$A_{\mathrm{lin}}$                 & 68.83 & [58.76, 78.86] & $\mathbf{-312.88}$ & 4/4 & does not transfer \\
perturbation sensitivity $\lambda$  & 2.96  & [0.65, 7.16]   & 0.908            & 3/4 & \textbf{recovers} \\
ours (gauge \texttt{G0\_RAW}, $\kappa=0$) & \textbf{0.28}  & [0.23, 4.86]   & $\mathbf{0.9989}$ & 4/4 & --- \\
\bottomrule
\end{tabular}
\end{table}

\paragraph{What the scorecard earns, and what it does not.} Our forecast dominates the recovering baseline
on every point estimate: mean absolute error $0.28$ and predictive $R^{2}$ $0.9989$,
both \textbf{at gauge} \texttt{G0\_RAW} ($\kappa=0$) and at no other, against $2.96$ and $0.908$
for the incumbent; every gauge the registered audit calls defensible puts $R^{2}_{\text{pred}}$
below zero, sign 4 of 4 against 3 of 4, and their error CIs \textbf{overlap heavily}
($[0.23,4.86]$ against $[0.65,7.16]$), so the two are not separable at $N=4$ under the identification
gate registered with the arms. It is the best
available forecaster on these data, and the design lacks the power to certify it against the incumbent. It
is not described as a validated predictor. The $A_{\mathrm{lin}}$ arm's predictive $R^{2}$ of $-312.88$ ($313\times$ worse than predicting the
roster mean) means a within-context steering-success feature carries \emph{anti-information} about a
cross-context horizontal shift. It \emph{does not transfer} to $\dec$; it does not mean the published
within-context result is wrong.

\subsection{The two figures that are withdrawn rather than nulled}
\label{app:fig:withdrawn}

The frozen figure plan registered two figures for the causal half of the paper (the measured movement of
$\dec$ under a scaffold edit, and the forecast against that movement) each with a pre-written null
render. \textbf{Neither may be drawn, and neither may be drawn on its null branch either.} A null render
communicates a measured null; these experiments produced no measurement at all, so drawing ``all CIs cross
zero'' would assert a measurement that does not exist. They are replaced by two figures that draw what was
actually measured: why the experiment was not launched.

\begin{figure}[htbp]
\centering
\includegraphics[width=0.9\textwidth]{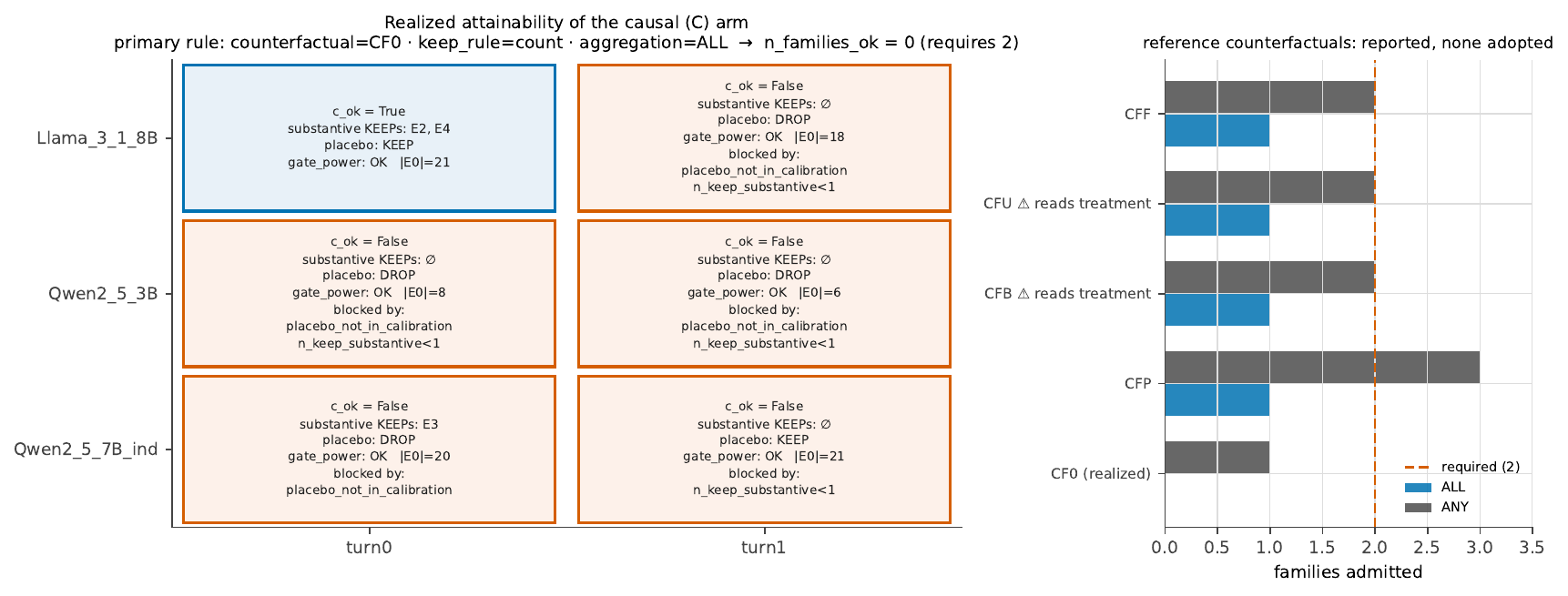}
\caption{\textbf{The attainability matrix: why the causal arm was not scorable.} The six realized
(family, readout-horizon) cells, each reporting the recoverability gate's substantive keeps, whether the
content-free placebo edit was itself kept (the in-calibration condition) the gate's power flag, and
the resulting cell verdict, beside four reference counterfactuals. The
derivation is \textbf{post-hoc}: it was written after all six cells and the scenario matrix existed, and
its own record says so. Read it as an exhaustive sensitivity analysis over a frozen roster, not as a
pre-data prediction. Encoding is
categorical: nothing here is signed. The power flag is printed in every cell so a reader can see the block
is not a power problem. Counterfactuals that read treatment information are marked as such and drawn
visually subordinate; the realized, frozen decision is the only one drawn as primary, and no
counterfactual column is presented as an alternative result.}
\label{fig:attainability}
\end{figure}

\begin{figure}[htbp]
\centering
\includegraphics[width=0.85\textwidth]{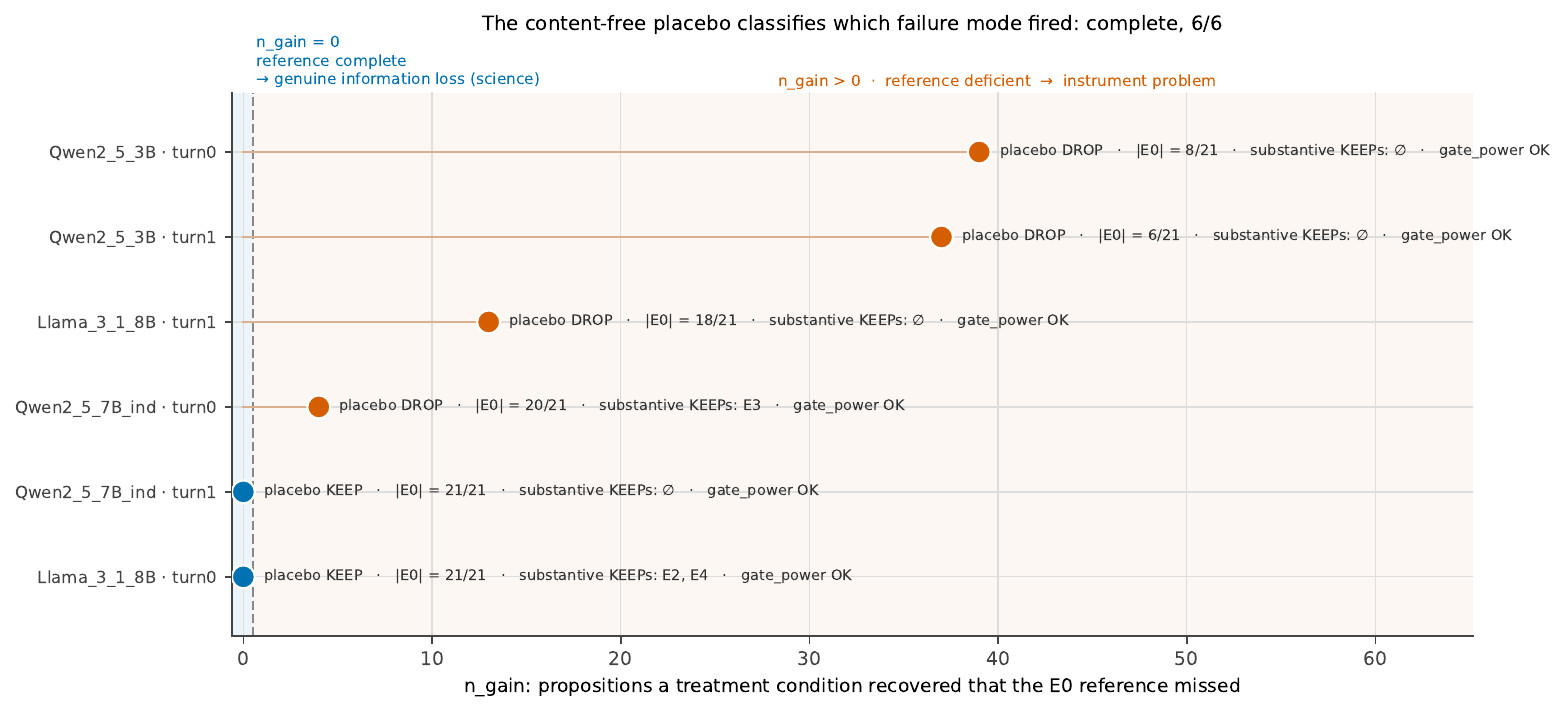}
\caption{\textbf{The placebo separates the gate's two failure modes on all six realized cells.} The same six cells
on two axes: the number of propositions the treatment conditions recovered that the reference read missed,
against the placebo's own gate decision. The separation is complete in 6 of 6 cells, which names which
failure mode fired in each: \emph{reference deficiency}, an instrument problem, against \emph{genuine
information loss}, a science result. The registered null render was an incomplete separation, drawn in the
off-diagonal quadrant with a caption saying the placebo is not a complete diagnostic; the layout does not
assume the result. The claim is a \emph{diagnostic} for which failure mode fired, never that the design was
validated.}
\label{fig:placebo}
\end{figure}

\section{Registered, and not run}
\label{app:notrun}

Three registrations in \S\ref{app:register} carry no result. All three are reported here rather than
deleted, because a registration that vanishes when its experiment does not happen is exactly what the
register-then-run discipline exists to prevent. \textbf{They are not the same kind of no-result}, and the
difference is stated rather than levelled: two were never launched, while the KV cache-reuse arm was run
in full and \emph{halted at its own registered gate}, which is a measurement about the arm and not an
absence of one.

\subsection{The causal scaffold intervention}
\label{app:notrun:scaffold}

The intervention that would establish $\dec$ as causally \emph{movable} (editing the ReAct format
scaffold with task information held fixed) was designed, pre-registered, priced at about \$205 of
agent-context generation, and \textbf{not launched}. Its free forward-pass gate returned zero rankable
cells in 6 of 6 (family $\times$ readout-horizon) cells, so the pre-registered planner refused to launch.
Budget and statistical power were both available; the missing thing was a readout that could rank the
edits.

\paragraph{The causal half alone was scored separately, and it is also unattainable.} The registered
headline splits into a causal claim and a forecast claim, and one registered branch for the realized
refutation outcome was to run \emph{the causal half only}, which uses no forecast. Its marginals are an
in-calibration placebo, at least one substantive keep, and a powered gate, on at least two families. Under
the registered reference and rule, no aggregation reaches two families. Four reference
repairs were scored, every one reported and none adopted: the repair is \textbf{necessary} (without it
no aggregation reaches two families) and \textbf{not sufficient} at the conservative rule, where every
repair and both keep rules return exactly one family. A reading of the gate restricted to the first
readout horizon would make the causal half attainable today on two families, unanimously across all repair
and rule combinations; \textbf{that reading is refused by name} in the registration's own analysis, because
choosing the horizon after seeing which horizon admits is the post-hoc selection the registration was
written to prevent.

\paragraph{What the refusal bought, and it is not nothing.} Two things, and both are reported. A
content-free placebo edit classifies the blocking gate's two failure modes in 6 of 6 cells, separating ``the
reference set is deficient'' from ``real information was lost'' (Fig.~\ref{fig:placebo}), which is what
makes this a diagnosis rather than a null. And an attainability calculus priced a headline experiment out
\emph{before} rather than after paying for it. The two near-miss cells fail for \emph{opposite} reasons,
and neither is a power problem.

\paragraph{Three threats retained verbatim and un-discharged} so that a future launch inherits them rather
than re-deriving them: the matched-information invariant is the crux and can only be audited, not proven;
the mechanism is a hypothesis and the experiment tests the prediction rather than the mechanism; and
holding the chat-context midpoint fixed by construction is a single point of failure, because an edit that
perturbs the chat leg at all moves the difference for a reason the design cannot see. A fourth constraint
is not a retained threat but a measured property of the edit space, and a future launch inherits it too:
ReAct scaffold edits are \emph{parser-bound}, since removing a structural marker changes the judged metric,
so the edit space is smaller than the design assumed.

\paragraph{The scoping statement, and its limit.} The readout invalidity is a property of the 2024-era
models on this roster; the current-cohort cell behaves differently. That is what makes this a scoping
statement rather than a dead end. It is also, precisely, a claim about a cohort of which we have one
member. The one format-scaffold manipulation this paper \emph{can} show is weaker than the registered one:
three doses, a different behaviour, and point estimates that were not blind. It is reported as a
demonstration that a scaffold edit moves the response at all, never as the registered intervention and never
as evidence for the forecast direction.

\subsection{The KV cache-reuse arm}
\label{app:notrun:cache}

\S\ref{app:m:defl} states the scope limit that our design excludes cache reuse by construction and has not
measured what cache reuse contributes. A registered arm converts that limit into a measurement: the same
agent leg on one cell under cache \emph{reuse}, against a recompute leg, with all outcome branches
(including the null) registered before any number exists. The registration proposed pairing the reuse leg
against the \emph{committed} recompute leg; what actually ran generated \emph{both} legs fresh under an
identical loop, and the paragraph below reports the realized design rather than the proposed one.

Because ``cache reuse'' is ambiguous between several implementable policies, and an ambiguous term silently
resolved after the data is exactly the failure register-then-run exists to prevent, a second registration
resolves the term to a named policy set \textbf{before} any machine is rented, and that resolution is itself
committed as a registration. \textbf{The arm then ran} (both legs newly generated under the identical
loop, $2\times7{,}400$ rollouts on one cell) and \textbf{halted at clause 1 of its registered $Q_0$
gate}: the two legs' item-id sets are $199$ and $200$ and the gate requires array equality, so the primary
questions ship as not-scored rather than as a cache-irrelevance finding the registration forbids reporting
from a failed gate. The manuscript's sentence therefore remains \emph{``absent by construction''} and
nothing stronger: not because the arm was skipped, but because it was run and its gate was honoured. Two registered branches on this study had no emitter when the study
landed; both estimators were then written, both are refits of records that already exist, and both shipped
the branch their pre-registration had named in advance (Appendix~\ref{app:registers}).

\subsection{Braun's geometric covariates}
\label{app:notrun:braun}

\S\ref{app:rel:braun} states the position: both covariates, in both deployment frames, on the pinned roster,
scored against $\dec$ under the same frozen leave-one-model-out rule: registered, and not measured. The
limitations section states the gap rather than the registration implying a result. The roster supports
$n=4$ models and the exact two-sided permutation floor at $n=4$ is $0.0833$, so the conditioning is
roster-limited by construction whenever it is run.

\section{The limitations register, in full}
\label{app:limits}

\subsection{Why this section, of all sections, needs a machine to score it}
\label{app:lim:L0}

The threat list opens with the reason, written before any paid number existed: \emph{a limitations section
written after the results is where a paper quietly drops whichever weaknesses its data happened to expose. A
reviewer cannot tell an honest limitation from a self-serving omission after the fact; the only defence is to
have enumerated the threats before knowing which ones the data would let us dodge.}

That defence was half-built. The enumeration happened (21 headings committed before the paid stage) but
nothing checked the enumeration against the paper. The corpus has since grown from 21 headings
to \textbf{66}, and the pre-register manuscript named \textbf{five} of them. A reader had no way to tell whether the other 61 were
absent because they were subsumed, because they were moot, or because they were inconvenient.

\textbf{So the register in \S\ref{app:lim:register} has the corpus as its denominator and this section as its
numerator.} Every heading in the frozen threat list gets exactly one row and exactly one disposition. A row
may not pass by being written down: the section it points at must exist and must carry that threat's marker
in its own body, so a pointer that goes nowhere is a defect rather than a skip. A threat present in the
frozen cut and dropped from the paper is not an ordinary defect but a named failure: the exact failure the
preamble above warned about.

\textbf{Two honest limits of the instrument itself.} Its denominator is our own threat list, so a threat
nobody in this project ever wrote down is invisible to it: it measures completeness \emph{relative to what we
already knew}, which is a real but bounded claim, and it is not a substitute for adversarial review. And a
threat can be discharged badly: the register checks that a threat is \emph{stated} where the row says it
is; it does not and cannot check that the statement is adequate.

\textbf{One disposition deserves its own warning.} \emph{Did not materialize} is the disposition most open to
abuse, because ``the data let us dodge it'' is exactly what a self-serving omission sounds like when written
down honestly. It is therefore the only disposition that requires an artifact pointer in its note, it is
confined to \S\ref{app:lim:L9}, and that section is written to say what the non-materialization \emph{cost}
rather than to bank it. There is exactly one such row, and it records a shortfall.

\subsection{Construct validity: what \texorpdfstring{$\dec$}{dEC50} reads, and what it does not}
\label{app:lim:L2}

The location is not a pure readout of the reparametrization: it reads the reparametrization together with the
baseline offset, which is an interpretive bound and the reason the paper's guardrails forbid claiming the
contexts differ \emph{only} horizontally. The 4PL can be misspecified, and one cell admitted a descending
fit. Coherence gating changes the denominator the estimator averages over, which is paired with the
coherence-as-response finding rather than treated as a nuisance. Equal upper plateaus mean the vertical half
of the reparametrization is \emph{unconstrained}, not shown absent. Coherence is fitted as a response throughout, and the reported flag is blind to its own
collapse. And the deliverable maximum is right-censored in four of
ten context-cells, so the efficacy contrast names its scale rather than being quoted scale-free.

\subsection{Identification of the fitted location: the grid, the interval, and the ceiling}
\label{app:lim:L3}

\paragraph{Two limitations of the positive anchor, stated here rather than only in the register.}
\textbf{(a)} The positive arm's displacement is read on a sub-grid: its \emph{sign} survives without a
curve model, but its magnitude is not resolved by these data, and no magnitude for it should be quoted
as settled. \textbf{(b)} The interval on that anchor is \emph{anticonservative}, which is why a repair
was registered separately rather than folded into the number. Both intervals are printed in
\S\ref{app:calib}: the committed one, which the registration dates and Table~\ref{tab:core} carries,
and the repaired one beside it, so a reader sees both rather than a silently improved figure.

The $\alpha^{*}$ ceiling is an estimate with its own uncertainty, and two registered families carry a null
value for it. The primary fit runs the full swept grid rather than the stricter registered dose exclusion,
and both readings are printed. The finer grid that resolves the positive arm fails the span half of the
identifiability rule, so the two grids are not pooled. And the visual stress test in the body is a
law-excluded cell: its second summary is not a corrected estimate, and its screen is post-hoc.

\subsection{The control spine as realized: four errata, and what would falsify each}
\label{app:lim:L4}

Appendix~\ref{app:errata} states the four errata and \S\ref{app:m:spine} states where each bites. The
register carries, in addition: scaffold integrity collapses with dose and $\alpha^{*}$ does not screen it, so
the top of the grid is the least trustworthy part of it; the judged object changes kind with dose; the judge
artifact and the estimand share a signature, which is measured and separated rather than assumed away; and
the one family with a valid gate readout is the one whose chat template folds the system role structurally.
Each erratum is stated with what would falsify it.

\subsection{Nuisance parameters, and the surfaces the readouts are taken on}
\label{app:lim:L5}

\paragraph{The readout surface, and where it reverses.} Every residual readout we \emph{report a
result from} is taken on the prompt, and the one study that read the alternative surface is reported
here rather than folded into those results. In the one cell where the comparison is movable, the frame difference \emph{reverses
sign} on the model's own generated tokens. We therefore make no claim about the generated-token
readout and scope the prompt-side result to the prompt.

The same commanded coefficient is injected at more sites in the agent context; matching the injection
phase set across the two contexts leaves the displacement intact and sign-identical, while the agent-only
sites do add potency, so a minority of it is dose bookkeeping and the majority is not. The carrier cannot be separated from the delimiters, so no residual-stream mechanism
is claimed. The localizing ladder runs on the leg that is \emph{not} information-matched and on the movable
metric, and the finest frame ladder is single-dose with an inherited ranking and a random band undefined in
17 of 20 cells.

\subsection{What the deflationary rejections license, and what they do not}
\label{app:lim:L6}

The norm null is rejected against one normalization model, on a turn-0 surface, under an assumption that
cannot be verified from this data. $\dec$ is not a translation along the steering direction, and the
neighbouring predictor does not transfer to it. The delivered-unit correction changes no sign, and the
transport model that would have explained the shift fails. Braun's covariates were claimed before being
computed, and the erratum is recorded rather than silently corrected. Every deflationary count is scored on
a five-cell panel frozen before the roster's sixth GREEN cell existed; the relative-dose commensurability
verdict is a five-cell result whose spread the sixth cell takes from $2.68$ to $21.74$, flipping it. A
post-hoc plateau-gap diagnostic correlated with the \emph{absolute} dose axis; scale-free, it never held:
on seven cells or on eight. And two consumers of a shared loader are invisible to the currency check because
its artifact set is keyed to a declaration; both are registered deferrals.

\subsection{The forecast: refuted out of sample, and the ways it could have been right for the wrong reason}
\label{app:lim:L7}

Parameter-free means no \emph{per-model} parameter, and the global choices are named rather than hidden. The
forecast shares its assumptions with the theory it is derived from, and the realized slope refutes its own
derivation. The arms did not fail benignly: the registered verdict is that a published baseline recovers,
and the forecast is never called validated. The readout is a point on a trajectory, the registered gauge
board reads gauge-dependent, and no entrant is promoted post hoc.

\subsection{The causal claim: threats retained un-discharged, because the experiment was not run}
\label{app:lim:L8}

\S\ref{app:notrun:scaffold} carries this group in full: three threats retained verbatim and un-discharged,
three that materialized as measurements and are reported as such, and one weaker manipulation that is
reported as a demonstration and never as the registered intervention.

\subsection{The one threat that did not materialize, and what its non-materialization cost}
\label{app:lim:L9}

Left-censored salvage bounds are directional rather than two-sided: where a fitted midpoint is only
\emph{bounded}, $\dec$ is a one-sided directional bound and weaker evidence than a two-sided CI, and the
frozen posture was that the salvage claim would never rest on a bound alone. \textbf{It did not arise}: no
cell in this paper is reported as a censored bound, so no headline rests on one.

\textbf{And that is not good news.} The reason no bound was reported is that the salvage arm produced no
salvage cell at all. The salvage question (whether a horizontal displacement is estimable in families
where a vertical gain ratio dies against a ceiling) was called, before the data, the cheapest thing that
could have made this paper, and \textbf{the shipped fit does not answer it}. The paid stage passes on
two-sidedness alone. The threat is moot because the arm it threatened is empty, which is a strictly worse
outcome than the threat materializing, and it is recorded as such rather than counted as a clean sheet.

\subsection{Statistical threats, external validity, and theory-dependence}
\label{app:lim:L10}

Multiplicity across families, arms and edits: exploratory verdicts are labelled as such and the law rests on
registered primaries. Greedy decoding gives item-resampling intervals with no decoding variance, which is a
scope statement rather than a defect. Per-sample variance is high by construction under the cylindrical
hypothesis, so the contrast is powered against variance rather than cell count (\S\ref{app:m:inference}).
The boundary of the claim across steering type, behaviour, roster, scaffold, decoding and language is stated
as a scope condition. And theory buys the motivation and the $\alpha^{*}$ scope condition: the difference of
fitted midpoints survives the theory failing.

\subsection{Operational and reproducibility threats, each one measured}
\label{app:lim:L11}

The guard suite had never been run as a suite, and a later audit found seven guards invariant to their
inputs. Most committed artifacts do not regenerate from the tree a reviewer receives, and every document
that quotes one names the tree it was measured in. The registered reading rules had no working call site
until a prereg-to-verdict ledger enforced them. One registered rule had no recorded verdict, and it was the
$\alpha^{*}$ control's own live/die criterion. Cells are pinned to checkpoints and one of the eight records what it
loaded, so run-time provenance is absent for the other seven. Each of these is printed at the size the
finding had, unflattering.

\subsection{The threat register}
\label{app:lim:register}

Every heading in the frozen threat list, with its disposition and the section that carries it.
\emph{container} rows are group headings whose children carry the content; \emph{scope} rows are boundary
statements rather than defects; \emph{moot (unrun)} rows are the causal-claim threats retained verbatim
against a future launch.

{\footnotesize
\setlength{\tabcolsep}{4pt}
\begin{longtable}{@{}p{0.05\textwidth}p{0.15\textwidth}p{0.075\textwidth}p{0.66\textwidth}@{}}
\caption{The threat register: every heading in the frozen threat list, its disposition, and the
section that carries it.}\label{tab:threats}\\
\toprule
threat & disposition & where & note \\
\midrule
\endfirsthead
\caption[]{The threat register \emph{(continued)}.}\\
\toprule
threat & disposition & where & note \\
\midrule
\endhead
1 & CONTAINER & \S\ref{app:lim:L2} & container heading for the construct-validity group; its four children carry the threat content \\
1.1 & APPENDIX & \S\ref{app:lim:L2} & the location reads the reparametrization together with the baseline offset; interpretive bound and guardrail G5 \\
1.2 & APPENDIX & \S\ref{app:lim:L2} & 4PL misspecification, its frozen mitigations, and the one cell where a descending fit was admitted \\
1.3 & APPENDIX & \S\ref{app:lim:L2} & coherence gating changes the denominator; paired with the coherence-as-response finding \\
1.4 & DID\_NOT\_MAT. & \S\ref{app:lim:L9} & no censored bound is reported in a committed artifact (bound\_direction null 8 of 8) because Q2\_salvage\_pass is false and Q2\_hits is empty; recorded as a shortfall \\
2 & CONTAINER & \S\ref{app:lim:L8} & container heading for the causal-claim group; framing only, children carry the threats \\
2.1 & MOOT\_UNRUN & \S\ref{app:lim:L8} & matched information can only be audited; retained verbatim per its own registration so a future launch inherits it \\
2.2 & MOOT\_UNRUN & \S\ref{app:lim:L8} & the experiment tests the prediction and not the mechanism; retained against its own registration \\
2.3 & MOOT\_UNRUN & \S\ref{app:lim:L8} & EC50\_chat held fixed is a single point of failure; retained against its own registration \\
2.4 & APPENDIX & \S\ref{app:lim:L8} & control and instrument want different readout points and the overlap is empty; this is the measurement that priced the launch out \\
2.5 & APPENDIX & \S\ref{app:lim:L8} & the two failures are marginal rather than joint and no horizon repairs them; cost was not the binding constraint \\
2.6 & APPENDIX & \S\ref{app:lim:L8} & the causal half alone fails on a different marginal, and the placebo classifies the gate 6 of 6 \\
3 & CONTAINER & \S\ref{app:lim:L7} & container heading for the predictor group; its four children carry the threat content \\
3.1 & APPENDIX & \S\ref{app:lim:L7} & parameter-free means no per-model parameter and the global choices are named \\
3.2 & APPENDIX & \S\ref{app:lim:L7} & the predictor shares its assumptions with the theory, and the realized slope refutes its own derivation \\
3.3 & APPENDIX & \S\ref{app:lim:L7} & the arms did not fail benignly; the registered verdict is STOP\_BASELINE\_RECOVERS and the predictor is never called validated \\
3.4 & APPENDIX & \S\ref{app:lim:L7} & the readout is a point on a trajectory, the gauge board reads GAUGE\_DEPENDENT, and no entrant is promoted post hoc \\
4 & APPENDIX & \S\ref{app:lim:L3} & the alpha-star ceiling is an estimate with its own uncertainty, and two registered families carry a null value \\
4.1 & APPENDIX & \S\ref{app:lim:L3} & the primary fit runs the full swept grid rather than the stricter registered dose exclusion; both readings printed \\
4.2 & APPENDIX & \S\ref{app:lim:L4} & Control A enters as a diagnostic and as the admission gate, never as a correction of the fitted response \\
4.3 & APPENDIX & \S\ref{app:lim:L4} & Control B matches the task and not one span of the system prompt; a bounded mismatch we name \\
4.4 & APPENDIX & \S\ref{app:lim:L4} & scaffold integrity collapses with dose and alpha-star does not screen it; the grid top is least trustworthy \\
4.5 & APPENDIX & \S\ref{app:lim:L4} & Control M is length-correlated rather than length-controlling, and the judged object changes kind with dose \\
4.6 & APPENDIX & \S\ref{app:lim:L4} & Control G ran on one clause of two; the beats-random clause was read by nothing \\
4.7 & APPENDIX & \S\ref{app:lim:L4} & Control N holds on the rollout record and not the fitted set, and the pairing is inert for a location estimand \\
4.8 & APPENDIX & \S\ref{app:lim:L11} & the guard suite had never been run as a suite, and a later audit found seven guards invariant to their inputs \\
4.9 & APPENDIX & \S\ref{app:lim:L11} & most committed artifacts do not regenerate from git archive HEAD; every quoting document names the tree \\
4.10 & APPENDIX & \S\ref{app:lim:L11} & the registered reading rules had no working call site; now enforced by the prereg-to-verdict ledger \\
4.11 & APPENDIX & \S\ref{app:lim:L11} & one registered rule had no recorded verdict and it was the alpha-star control's own live-die \\
4.12 & APPENDIX & \S\ref{app:lim:L4} & the judge artifact and the estimand share a signature; measured and separated rather than assumed away \\
4.13 & APPENDIX & \S\ref{app:lim:L3} & the positive arm is a sub-grid displacement whose sign survives without a curve model and whose magnitude is unresolved \\
4.14 & APPENDIX & \S\ref{app:lim:L3} & the interval is anticonservative on the positive anchor and armed a separately registered repair rather than a number change \\
4.15 & APPENDIX & \S\ref{app:lim:L3} & the finer grid that resolves the positive arm fails the span half of the identifiability rule; the two are not pooled \\
4.16 & APPENDIX & \S\ref{app:lim:L11} & cells are pinned to checkpoints and 1 of 8 records what it loaded; run-time provenance is absent for seven of the eight fitted cells \\
4.17 & BODY & \S\ref{sec:disc} & a dose is not a shared unit, so cross-family magnitudes carry a normalization and the log-ratio is the poolable companion \\
4.18 & BODY & \S\ref{sec:disc} & the curves are not translates; PI\_Q2\_TRANSLATION\_REJECTED, and dEC50 is a location summary at one response level \\
4.19 & APPENDIX & \S\ref{app:lim:L2} & equal plateaus mean the vertical half of the reparametrization is unconstrained rather than shown absent \\
4.20 & APPENDIX & \S\ref{app:lim:L2} & coherence is a response not a filter, and the reported flag is blind to its own collapse \\
4.21 & APPENDIX & \S\ref{app:lim:L6} & the gain is dose-indexed and two of our own registered predictions about it were refuted in Results \\
4.22 & APPENDIX & \S\ref{app:lim:L2} & the deliverable maximum is right-censored in four of ten context-cells and the efficacy contrast names its scale \\
4.23 & APPENDIX & \S\ref{app:lim:L6} & the norm null is rejected against one normalization model on a turn-0 surface with an unverifiable assumption \\
4.24 & APPENDIX & \S\ref{app:lim:L4} & the one family with a valid gate readout is the one whose chat template folds the system role structurally \\
4.25 & BODY & \S\ref{sec:disc} & the injection layer is one per family and was chosen on the chat leg for the Qwen anchor; still unswept \\
4.26 & BODY & \S\ref{sec:disc} & the direction is chat-extracted and unvaried in the reported fits; swept only on already-public reciprocal data \\
4.27 & APPENDIX & \S\ref{app:lim:L5} & the same commanded alpha is injected at more sites in the agent context; matching the phase set leaves the displacement intact and a minority of it is dose bookkeeping \\
4.28 & APPENDIX & \S\ref{app:lim:L6} & dEC50 is not a translation along the steering direction, and the neighbouring predictor does not transfer to it \\
4.29 & APPENDIX & \S\ref{app:lim:L6} & Braun's covariates were claimed before being computed; the erratum is recorded rather than silently corrected \\
4.30 & APPENDIX & \S\ref{app:lim:L8} & the showable scaffold manipulation is three doses in another behaviour with non-blind point estimates \\
4.31 & APPENDIX & \S\ref{app:lim:L5} & the carrier cannot be separated from the delimiters, so no residual-stream mechanism is claimed \\
4.32 & APPENDIX & \S\ref{app:lim:L6} & the delivered-unit correction changes no sign and the transport model that would have explained the shift fails \\
4.33 & APPENDIX & \S\ref{app:lim:L5} & every residual readout is on the prompt and the frame difference reverses on generated tokens in the one movable cell \\
4.34 & APPENDIX & \S\ref{app:lim:L5} & the localizing ladder runs on the leg that is not information-matched and on the movable metric \\
4.35 & APPENDIX & \S\ref{app:lim:L5} & the finest frame ladder is single-dose with an inherited ranking and a random band undefined in 17 of 20 cells \\
4.36 & APPENDIX & \S\ref{app:lim:L3} & the visual stress test is a law-excluded cell; its second summary is not a corrected estimate and the screen is post-hoc \\
4.37 & APPENDIX & \S\ref{app:lim:L6} & every deflationary-accounts count is scored on a five-cell panel frozen before the roster's sixth $\alpha^{*}$-GREEN cell existed \\
4.38 & APPENDIX & \S\ref{app:lim:L6} & relative-dose commensurability (NR-Q6) is a five-cell result; the sixth $\alpha^{*}$-GREEN cell takes the spread from 2.68 to 21.74 and flips the verdict \\
4.39 & APPENDIX & \S\ref{app:lim:L6} & the post-hoc plateau-gap diagnostic correlated with the absolute dose axis; scale-free it never held, on seven cells or eight \\
4.40 & APPENDIX & \S\ref{app:lim:L6} & roster\_currency's artifact set is keyed to a declaration, so two consumers of the shared loader are invisible to it; both are registered deferrals \\
4.41 & APPENDIX & \S\ref{app:lim:L6} & a deliberately pinned historical record scored as stale; the exemption that fixes it is measured on four clauses, never listed \\
5 & CONTAINER & \S\ref{app:lim:L10} & container heading for the statistical group; its three children carry the threat content \\
5.1 & APPENDIX & \S\ref{app:lim:L10} & multiplicity across families arms and edits; exploratory verdicts are labelled and the law rests on registered primaries \\
5.2 & SCOPE & \S\ref{app:lim:L10} & greedy decoding gives item-resampling intervals with no decoding variance; a scope statement rather than a defect \\
5.3 & APPENDIX & \S\ref{app:lim:L10} & per-sample variance is high by construction under CRH, so the contrast is powered against variance not cell count \\
6 & SCOPE & \S\ref{app:lim:L10} & the boundary of the claim across steering type behaviour roster scaffold decoding and language \\
7 & SCOPE & \S\ref{app:lim:L10} & theory buys the motivation and the alpha-star scope condition; the difference of fitted midpoints survives the theory failing \\
8 & APPENDIX & \S\ref{app:lim:L11} & operational disciplines were measured rather than asserted, and the measurements are printed unflattering \\
\bottomrule
\end{longtable}
}

\paragraph{What a reader should take from this register.} Not that the paper is unusually fragile: it is
that the enumeration was frozen before the data and then scored against the manuscript by a machine, so the
list a reviewer sees is the list that existed before we knew which threats the data would let us dodge. The
instrument's denominator is our own threat list, which is the bound stated in \S\ref{app:lim:L0} and is not
a substitute for adversarial review.

\bibliographystyle{iclr2026_conference}
\bibliography{refs,refs_named}

\ifanonymous\else

\section*{Errata to the public record of the predecessor preprint}
\label{app:predecessor-errata}

This version replaces \texttt{arXiv:2607.09156v1} in place. That
version was never revised, so the corrections below were live in the public record until now, and
they are stated here rather than repaired in silence: a replacement that quietly fixed them would
leave no trace that the earlier reading had been published and relied upon. Each was checked against the published PDF; none of them moves a number, an interval or a
verdict in the present paper, which re-fits its estimand from the raw rollouts.

\begin{enumerate}\itemsep2pt

\item \textbf{``No universal sign'' is wrong; attenuation is not inversion.} The predecessor's
contribution list, limitations and conclusion each state that the coupling distribution establishes
``no universal coupling constant and, via one clean attenuator, no universal sign.'' Every family's
coupling estimate is positive ($2.00$, $1.50$, $1.45$, $0.93$, $0.78$, $0.43$), and a coupling of
$0.43$ attenuates the chat effect without inverting it. The distribution has one sign. The
\emph{no universal constant} half is unaffected and stands.

\item \textbf{The cross-behavior contrast is arithmetically wrong, and $0.57$ is a different
estimator.} The predecessor reads ``refusal bypass at $1.45$ versus sycophancy at $0.78$ yields a
difference of $0.57$''. The difference of those two estimates is $0.67$. The quoted $0.57$, and the
interval $[0.06, 0.99]$ reported with it, belong to a k-ratio contrast computed on a different
quantity, not to the subtraction the sentence describes.

\item \textbf{One extraction recipe is stated for two behaviors that used different ones.} The
$128$-versus-$128$ last-token difference-of-means construction is the refusal recipe. The sycophancy
direction follows the persona-vectors construction, averaged over response positions rather than read
at the last token.

\item \textbf{The $0.88$ phase-localization share has the wrong denominator.} Pre-observation block
injection reproduces $0.88$ of the full agent \emph{effect} ($42.0$ of $47.5$ points), not of the
coupling gain; against the agent-over-chat excess of $11.5$ points the share is roughly half. The
localization conclusion is unchanged: pre-observation carries the effect and post-observation
returns $0.00$, and only the denominator in that sentence is wrong.

\item \textbf{A prior-work result is attributed to the wrong model scale.} The predecessor describes
\citet{lermen2024ablation} as transferring refusal-direction ablation to Llama-3.1-8B agents. That
work reports Llama-3.1-70B agents; the predecessor's own reference list carries the correct title.

\item \textbf{\citet{tan2024generalisation} is under-cited relative to the overlap.} The predecessor
mentions that work once, as documenting out-of-distribution brittleness of steering in chat. It does
more that bears directly on the claim: it varies the evaluation context and reports the resulting
change in effect, and it defines \emph{relative steerability}, a ratio of fitted slopes across
conditions. The transfer ratio $T$ belongs to that family of object and should have been positioned as
an instance of it measured on a new axis. The present paper's contribution is not the ratio but its
non-identification, and the horizontal parameter that replaces it.

\item \textbf{``Specific to the additive mechanism'' claims more than the design delivers.} The
additive and ablation arms were run at their closest tested doses, with chat swings of $32.3$ and
$56.0$ points. The contrast is between intervention families at unmatched effect size, not an
effect-matched identification of a mechanism. The $20.1$-point gain difference and its interval stand
as reported; the word \emph{specific} does not.

\item \textbf{The uniform-protocol sentence overstates uniformity on two counts, and its gate clause
was scored under a different metric.} Yi-1.5-9B analyses $183$ paired items, not the stated $200$; one
Qwen2-7B cell ran four matched-norm random directions, not five. Separately, the sub-saturation gate
was applied at each family's chat pilot under a backstop-parser metric, while the reported primary is
the length-controlled combined binary: the metric changed underneath the selection. We disclose the
selection metric rather than re-running selection under the primary, which would let dose choice see
the coupling outcome.

\end{enumerate}

\noindent Two candidate errata were checked and are \emph{not} corrections to that version, and are
named here so the list is not read as exhaustive by omission: its statement that three roster families
yield no coupling estimate is correct for the roster it ships, and the descriptions of directional
ablation as a weight-matrix edit and the malformed control sentence both belong to a later cut and do
not occur in the published text.

\fi

\end{document}